\documentclass{article}
\usepackage[accepted]{ICML2023_format/icml2023}
\usepackage{color}

\NewDocumentCommand{\var}{O{s} m O{}}{%
  \ensuremath{#1_{#2}^{#3}}
}
\usepackage{siunitx}

\newcommand{\commentout}[1]{}

\definecolor{light-gray}{gray}{0.80}

\newcommand\appref{Appendix~\ref}

\newcommand\fref{Figure~\ref}
\newcommand\tref{Table~\ref}
\newcommand\sref{Section~\ref}

\usepackage{amsthm}

\usepackage{amsthm}

\newtheorem{mytheorem}{Theorem}
\newtheorem{assumption}[mytheorem]{Assumption}
\newtheorem{lemma}[mytheorem]{Lemma}
\newtheorem{remk}[mytheorem]{Remark}

\usepackage{amsmath,amsfonts,bm}

\def\eqref#1{equation~\ref{#1}}

\def\1{\bm{1}}

\def\rvx{{\mathbf{x}}}

\def\rmW{{\mathbf{W}}}

\DeclareMathAlphabet{\mathsfit}{\encodingdefault}{\sfdefault}{m}{sl}
\SetMathAlphabet{\mathsfit}{bold}{\encodingdefault}{\sfdefault}{bx}{n}

\def\sR{{\mathbb{R}}}

\usepackage{lipsum,booktabs}
\usepackage{multirow}
\usepackage{booktabs}
\usepackage{nopageno}
\usepackage{booktabs}
\usepackage{natbib}
\usepackage{hyperref}
\definecolor{myblue}{rgb}{0,0.2,0.8}
\usepackage{lipsum}
\makeatletter
\def\adl@drawiv#1#2#3{%
        \hskip.5\tabcolsep
        \xleaders#3{#2.5\@tempdimb #1{1}#2.5\@tempdimb}%
                #2\z@ plus1fil minus1fil\relax
        \hskip.5\tabcolsep}
\newcommand{\cdashlinelr}[1]{%
  \noalign{\vskip\aboverulesep
           \global\let\@dashdrawstore\adl@draw
           \global\let\adl@draw\adl@drawiv}
  \cdashline{#1}
  \noalign{\global\let\adl@draw\@dashdrawstore
           \vskip\belowrulesep}}
\makeatother

\definecolor{upforestgreen}{rgb}{0.6, 0.8, 0.2}

\usepackage{subcaption}
\usepackage{graphicx}
\usepackage{caption}
\usepackage{wrapfig}
\usepackage{chngcntr}
\usepackage{adjustbox}
\usepackage{arydshln}

\usepackage{tcolorbox}

\icmltitlerunning{Submission and Formatting Instructions for ICML 2023}

\begin{document}

\twocolumn[
\icmltitle{Understanding INT4 Quantization for Language Models: \\Latency Speedup, Composability, and Failure Cases}



\icmlsetsymbol{equal}{*}
\begin{icmlauthorlist}
\icmlauthor{Xiaoxia Wu}{equal,yyy}
\icmlauthor{Cheng Li}{equal,yyy}
\icmlauthor{Reza Yazdani Aminabadi}{yyy}
\icmlauthor{Zhewei Yao}{yyy}
\icmlauthor{Yuxiong He}{yyy}
\end{icmlauthorlist}


\icmlcorrespondingauthor{Xiaoxia Wu}{xiaoxiawu@microsoft.com}
\icmlcorrespondingauthor{Cheng Li}{chengli1@microsoft.com}
\icmlaffiliation{yyy}{Deepspeed Team @ Microsoft. Code is released at \url{https://github.com/microsoft/DeepSpeed}}
\icmlkeywords{Machine Learning, ICML}

\vskip 0.3in
]



\printAffiliationsAndNotice{\icmlEqualContribution} 

\begin{abstract}
Improving the deployment efficiency of transformer-based language models has been challenging given their high computation and memory cost. 
While INT8 quantization has recently been shown to be effective in reducing both the memory cost and latency while preserving model accuracy, it remains unclear whether we can leverage INT4 (which doubles peak hardware throughput) to achieve further latency improvement.  In this study, we explore the feasibility of employing INT4 weight and activation (W4A4) quantization for language models. Our findings indicate that W4A4 quantization introduces no to negligible accuracy degradation for encoder-only and encoder-decoder models, but causes a significant accuracy drop for decoder-only models.  
To materialize the performance gain using W4A4, we develop a highly-optimized end-to-end W4A4 encoder inference pipeline supporting different quantization strategies.
Our INT4 pipeline is $8.5\times$ faster for latency-oriented scenarios and up to $3\times$ for throughput-oriented scenarios compared to the inference of FP16, and improves the SOTA BERT INT8 performance from FasterTransformer by up to $1.7\times$.
We provide insights into the failure cases when applying W4A4 to decoder-only models, and further explore the compatibility of INT4 quantization with other compression methods, like pruning and layer reduction.
\end{abstract}

\section{Introduction}

As pre-trained large language models (LLMs) \citep{vaswani2017attention} such as BERT~\citep{tenney2019bert}, BART~\cite{lewis2020bart}, and GPT~\citep{radford2019gpt} require a significant amount of GPU resources to deploy, compression becomes a common practice to optimize model inference, especially for resource-constrained environments. 
One of the widely used compression techniques is quantization where data are stored and manipulated in a lower-precision format, such as 8-bit or 4-bit integers instead of 32-bit or 16-bit floating-point numbers.
It not only reduces the amount of memory required to store the model, but also can leverage the higher GEMM computation throughput for lower-bit data types on supported GPUs (e.g., peak INT4 Tensor Core TFLOPS doubles that of INT8 and quadruples that of FP16) to improve inference latency. 
Note that only quantizing the model weights without computing in lower-bit data types (i.e., keeping activation in FP16 or FP32) introduces no latency improvement (or even slower due to type conversion at runtime) but only memory saving.

Recent work proposes techniques to apply INT8 quantization (using INT8 computation where both weight and activation are quantized, referred to as W8A8) to all linear layers without introducing accuracy degradation for transformers
\citep{yao2022zeroquant,xiao2022smoothquant,dettmers2022gptint, dettmers2022llm,li2022dq,kim2021bert}.
\citet{yao2022zeroquant} also present an INT8 inference pipeline and show good end-to-end (E2E) performance improvement over FP16 model inference.
NVIDIA's FasterTransformer~\citep{fastertransformer} holds SOTA open-source INT8 implementations  where aggressive quantization are explored: mode-1 quantizes the attention computation beyond linear layers, and mode-2 further quantizes the residual connection trading off accuracy for latency.

 While we are advancing W8A8 quantization algorithms and implementations proven to be effective for LLMs, the questions arise: (1) whether INT4 inference (using INT4 computation where both activation and weight are quantized, referred to as W4A4) is feasible (acceptable accuracy drop) for these models, and (2) how it can be leveraged for performance improvement on real hardware.
 Although W4A4 has been successfully applied to other model types or hardware, e.g., convolution models for image classification with quantization-aware training strategy (QAT)~\citep{abdolrashidi2021pareto},\footnote{QAT requires the full training pipeline by quantizing the weight and activation during the forward process and updating the weights with gradients computed by straight through estimator \citep{bengio2013estimating} or other methods.} there is lack of work on exploring W4A4 for LLMs inference on GPU. 
 \citet{dettmers2022case} show little accuracy loss for LLMs when only model weights are quantized to 4-bit with post-quantization training (PTQ)\footnote{PTQ means the quantized model is arrived directly by mapping the weights from floating-point to low precision values without the full pipeline training (dataset and backward gradient).}, while the computation is still in FP16 as the activations are not quantized. 
\citet{wu2022extreme} prove that even the binary network can result in only a small degradation if applying QAT with knowledge distillation (KD)~\citep{hinton2015distilling} and longer training, but the activations are quantized to INT8 (using INT8 computation, not INT4). 
\citet{tang2022mkq} are the first to claim to apply W4A4 to BERT for inference with QAT and KD. However, their quantization method fails to enable W4A4 for all but only the last two layers in a four-layer TinyBERT model (otherwise causing drastic accuracy drops). Moreover, their E2E INT4 inference lacks implementation details, with conflicting performance numbers when compared to FasterTransformer (see \appref{sec:mkq-results}).
  
In this work, we aim not only to better understand the accuracy impact of INT4 quantization on common LLMs, but also to materialize and maximize the benefit of using INT4 computation in E2E inference, further improving the SOTA inference performance on LLMs.
Specifically, we make the following contributions:
 \begin{itemize}
     \item We explore the feasibility of W4A4 quantization across popular language model types, by leveraging the recent layer-wise knowledge distillation method for quantization. We show that our W4A4 can achieve no accuracy loss for the encoder-only models (BERT) on classification problems, negligible accuracy difference for encoder-decoder models (BART) on summarization tasks, but causes a relatively larger accuracy drop for decoder-only models (GPT) on autoregressive generation tasks.
     
     \item We develop a highly optimized end-to-end encoder model inference pipeline to support INT4 computation. The pipeline is built with modular components supporting different quantization strategies to accommodate latency- or throughput-oriented scenarios. 
     Our inference pipeline is up to $8.5\times$/$3\times$ faster for latency-/throughput-oriented scenarios when compared to HuggingFace FP16 BERT implementation, and improves the SOTA BERT INT8 performance from NVIDIA FasterTransformer by up to $1.7\times$.
        
    \item To unveil the causes of larger accuracy drop for decoder-only models (GPT) when using INT4 quantization, we provide an in-depth analysis of layer normalization, pretraining effect, and attention mechanism. 
    Additionally, we study the composability of INT4 quantization with other compression techniques, including pruning and layer-reduction, for encoder-related models. 
  
 \end{itemize}


\section{Related Work}
\label{sec:related_work}
 
 Model compression, as a technique to reduce to the model size and computation costs, can be achieved by pruning, quantization, low-rank factorization and efficient architecture designs~\citep{han2015learning, li2016pruning,mao2017exploring,lecun1990optimal,michel2019sixteen, fan2019reducing, gordon2020compressing, raganato2020fixed,dong2019hawq,yao2021mlpruning, mao2020ladabert,hinton2015distilling,sanh2019distilbert, sun2019patient,jiao2019tinybert, sun2020mobilebert, wang2020minilm, lan2019albert, dehghani2018universal,liu2021post,hu2021lora,micikevicius2018mixed,polino2018model,frantar2022optimal}.
 Among the large body of litterateurs,  we mainly cover the recent related works on INT4 quantization and system inference. 

As described in the introduction, the 8-bit quantization for LLMs, and/or mixing with other precision, has been widely studied and proven to be effective in recent years \citep{yao2022zeroquant,xiao2022smoothquant,dettmers2022gptint, dettmers2022llm,li2022dq,frantar2022gptq,kim2021bert}. However, the purely INT4 quantization, as a very aggressive technique that can have a significant impact on the accuracy of the model, is not widely used in practice and  still emerging. To the best of our knowledge, we describe some more closely related works besides those mentioned in the introduction. In \cite{sun2020ultra}, a 4-bit floating point format with an adaptive gradient scaling technique is proposed to demonstrate its effectiveness in computer vision, speech and NLP tasks and solid hardware acceleration. Our study focuses on the use of INT4 quantization instead of FP4 and the acceleration hardware is based on the Ampere structure. In \cite{chung2020extremely}, a low-bits mixed precision quantization strategy is proposed to represent Transformer models. However, their activations are kept in full precision. In \cite{han2020convolutional}, a detailed implementation of INT4 optimization is presented, but it is only applicable to convolution networks and not transformer models. \cite{dettmers2022case,yao2023comprehensive,frantar2022gptq} study the INT4 weight quantization for transformers but the activation is not INT4 but FP16 or INT8, and they mainly focus on post-training quantization.

\section{Model Accuracy for INT4 Quantization}
\subsection{Quantization Algorithms and Training}\label{sec:methodology}




\textbf{Quantization.} For completeness, we here explain the symmetric and asymmetric quantization algorithms~\citep{yao2022zeroquant}. Suppose $\rvx\in \mathbb{R}^d$ and $\rvx_{q}\in \mathbb{R}^d$ represent respectively a full-precision and a quantized vector. The 
uniform symmetric mapping strategy from $\rvx$ and $\rvx_{int}$ is
\begin{equation*}
\small
    \rvx_{q}^{(sym)} = S\left\lceil clamp(\rvx/S; -2^{b-1}, 2^{b-1}-1)\right\rceil,
\end{equation*}
where $clamp$  restricts the value of its argument to a given range from $-2^{b-1}$ to $2^{b-1}-1$, 
$b$ is the number of bits used to represent the quantized value, $\lceil\cdot\rceil$ is the rounding operator, and $S \in \mathbb{R}$ is the scaling factor. 
For example, $S$ can be  computed as the maximum of the absolute elements in the  $\rvx$ vector, i.e.,
$S = max\left( abs(\rvx) \right)$.  
On the other hand,   the asymmetric mapping strategy can be expressed as
\begin{equation*}
\small
    \rvx_{q}^{(asym)} = S\left\lceil clamp( (\rvx-x_{\text{zero}}\mathbf{1})/S; 0, 2^{b-1}-1)\right\rceil +x_{\text{zero}} \mathbf{1},
\end{equation*}
where $\rvx_{\text{zero}}$  is used as a reference point  potentially reducing any bias into the asymmetric vector. The scalar $S$ can be computed as
$S = max(\rvx)-min(\rvx)$ and $x_{\text{zero}}=min(\rvx)$.

Throughout the paper, we always do both weight and activation quantization using the method proposed in~\citet{yao2022zeroquant}. See \appref{sec:method-app} for more details.

\textbf{Knowledge Distillation.}  Knowledge distillation (KD) can greatly improve the performance of quantized transformer models. 
It trains a smaller quantized model (the student model) by incorporating the knowledge from the larger full-precision model (the teacher model).  
This can be done by training the student model to mimic the behavior of the teacher model on the training dataset, using the output probabilities as a soft target~\citep{hinton2015distilling} and the hidden states (and/or attention maps) of each transformer layer to align feature maps~\citep{jiao2019tinybert,wang2020minilm,bai2020binarybert,li2016ternary,wu2022extreme}. 






\begin{table*}[ht]
\caption{The best quality for BERT/BART/GPT-type models (two sizes) over the validation datasets, respectively with metric Accuracy (Acc., higher is better), Rouge Lsum (RLsum, higher is better), and perplexity (PPL, lower is better). 
}\label{table:main-results}
\begin{adjustbox}{width=0.999\linewidth}
\centering
\begin{tabular}{l|cc|cc|ccc}
\toprule
Models     & \multicolumn{2}{c|}{BERT-base  (110M)}                         & \multicolumn{2}{c|}{BART-base (140M)}                                  & \multicolumn{3}{c}{GPT2-base (117M)}                                         \\
Tasks      & \multicolumn{1}{c}{MNLI-m/mm} & \multicolumn{1}{c|}{QQP}    & \multicolumn{1}{c}{CNNDailyMail} & \multicolumn{1}{c|}{XSUM}     & \multicolumn{1}{c}{PTB} & \multicolumn{1}{c}{WIKI-2} & \multicolumn{1}{c}{WIKI-103}\\
Metrics    & \multicolumn{1}{c}{Acc/Acc}  & \multicolumn{1}{c|}{F1/Acc} & \multicolumn{1}{c}{R1/R2/RLsum}     & \multicolumn{1}{c|}{R1/R2/RL} & \multicolumn{1}{c}{Perplexity} & \multicolumn{1}{c}{Perplexity}    & \multicolumn{1}{c}{Perplexity}       \\    \midrule  
FP32 (teacher) & 84.20/84.67 & 87.83/90.95 & 45.62/22.85/42.87 & 42.18/19.44/34.36 & 19.31 & 21.02  & 17.46\\
W4A4 (symmetric) &  84.31/84.48 & 88.11/91.14 & 44.63/21.42/41.92 & 41.54/18.61/33.69 & 22.17 &27.28 & 21.75\\
W4A4 (asymmetric)& 84.29/84.65 & 88.17/91.19 &    44.83/21.67/42.08                & 41.53/18.56/33.62 & 21.72 & 25.99   &21.54                                    \\
\end{tabular}
\end{adjustbox}
\begin{adjustbox}{width=0.999\linewidth}
\centering
\begin{tabular}{l|cc|cc|ccc}
\toprule
Models     & \multicolumn{2}{c|}{BERT-large  (345M)}                         & \multicolumn{2}{c|}{BART-large (406M)}                                  & \multicolumn{3}{c}{GPT2-medium  (355M) }                                         \\
Tasks      & \multicolumn{1}{c}{MNLI-m/mm} & \multicolumn{1}{c|}{QQP}    & \multicolumn{1}{c}{CNNDailyMail} & \multicolumn{1}{c|}{XSUM}     & \multicolumn{1}{c}{PTB} & \multicolumn{1}{c}{WIKI-2} & \multicolumn{1}{c}{WIKI-103}\\
Metrics    & \multicolumn{1}{c}{Acc/Acc}  & \multicolumn{1}{c|}{F1/Acc} & \multicolumn{1}{c}{R1/R2/RLsum}     & \multicolumn{1}{c|}{R1/R2/RL} & \multicolumn{1}{c}{Perplexity} & \multicolumn{1}{c}{Perplexity}    & \multicolumn{1}{c}{Perplexity}       \\   \midrule  
FP32 (teacher) &  86.65/85.91 & 88.08/91.07 & 44.82/21.67/41.80 & 45.42/22.37/37.29 & 15.92 & 15.92 &12.75 \\
W4A4 (symmetric)  & 86.25/86.20  & 88.30/91.17 & 45.12/21.73/42.31  & 44.39/21.28/36.33 & 17.69 & 19.51 &  14.57\\
W4A4 (asymmetric)&  86.49/86.28 & 88.35/91.24 &  45.20/21.85/42.40  & 44.91/21.74/36.79 & 17.32 & 18.74  &     14.23                                 \\
 \bottomrule
\end{tabular}
\end{adjustbox}
\end{table*}

\subsection{INT4 Quantization for Language Models}
\label{sec:results}
We perform the 4-bit quantization on all linear layers using QAT and KD.
We use BERT-base and BERT-large \citep{tenney2019bert} as representatives for encoder-only models and fine-tune them on two largest GLUE tasks, i.e., QQP~\citep{iyer2017first} and MNLI~\citep{williams2017broad} for small accuracy variations. 
We use GPT2 and GPT2-medium \citep{radford2019gpt} as representatives for decoder-only models and fine-tune them on three causal generation tasks, i.e., PTB~\citep{marcinkiewicz1994building}, 
Wikitext-2, and Wikitext-103~\citep{merity2016pointer}.
Finally, we use BART-base and BART-large as representatives for encoder-decoder models and fine-tune them on two summarization tasks, i.e., CNNDaiyMail~\citep{hermann2015teaching}, 
and XSum~\citep{narayan2018don}. 
In order to reduce the hyper-parameters' effect, e.g., the best quantization configuration for BERT may be suboptimal for GPT,
we exhaustively search hyper-parameters including iterations, learning rate, dropout, quantization groups, clip values, and knowledge distillation terms for each model and choose the best one to report here. 
We include the experimental details in~\appref{sec:experiment-setup} and~\tref{table:main-results-hp}.

We present the main results in~\tref{table:main-results} for both symmetric and asymmetric quantizations. 
We also provide more detailed iterative-vs-accuracy plots in~\fref{fig:main-results} on the validation datasets for QAT.
For symmetric quantization, as can be seen, there is no accuracy degradation for BERT models and negligible drops ($\le1$ point) for BART models, while the 4-bit decoder models, i.e., GPT2 and GPT2-medium, show a significant drop in perplexity ($\ge1.5$ points) compared to the original FP32 models.  
This suggests that classification/summarization tasks using encoder-only/encoder-decoder models are much more robust to quantization when compared to auto-regressive generation tasks using decoder-only models. 

Asymmetric quantization generally improves the accuracy performance over symmetric quantization since it better utilizes the quantization range. 
One notable thing is that even with a better quantization scheme (i.e., asymmetric quantization) and exhaustive hyper-parameter tuning, decoder-only models still have larger quality degradation compared to encoder-only and encoder-decoder models. 
To provide more insight into why decoder-only models are more sensitive to INT4 quantization, we give a detailed analysis in~\sref{sec:results-gpt}.

\section{Highly Optimized INT4 Encoder Inference }
\label{sec:system-perf}
To materialize and maximize the benefits of using INT4 computation in model inference, we develop a set of custom GPU kernels and an E2E highly optimized pipeline to support inference with INT4 (as well as INT8) quantized encoder models. 
We adopt the system optimizations described in~\citep{yao2022zeroquant} and~\citep{aminabadi2022deepspeed} when applicable, and take advantage of FlashAttention~\citep{dao2022flashattention} and the CUDA graph~\citep{cuda-graph} to further improve the performance. Moreover, we explore different quantization strategies for latency- or throughput-oriented 
 scenarios. The software design and implementation also largely apply to other model types, e.g., GPT decoders if the accuracy drop can be resolved.

We conduct the performance experiments on a Lambda A6000 workstation~\cite{lambda-box} ($2\times$A6000-$48$GB-GPU, $256$GB DRAM, and $2$TB NVME), with the following software setup: HuggingFace transformers 4.25.1, NVIDIA FasterTransformer v5.2.1,  PyTorch: 1.12.1, cuda 11.7, and cutlass v2.6.0. Currently, INT4 GEMM is not supported by CUBLAS, and is only available through CUTLASS~\citep{cutlass} and we use that to support the INT4 computation in model inference.

\begin{figure*}[t]
\begin{subfigure}{0.5\textwidth} {\includegraphics[width=1\linewidth]{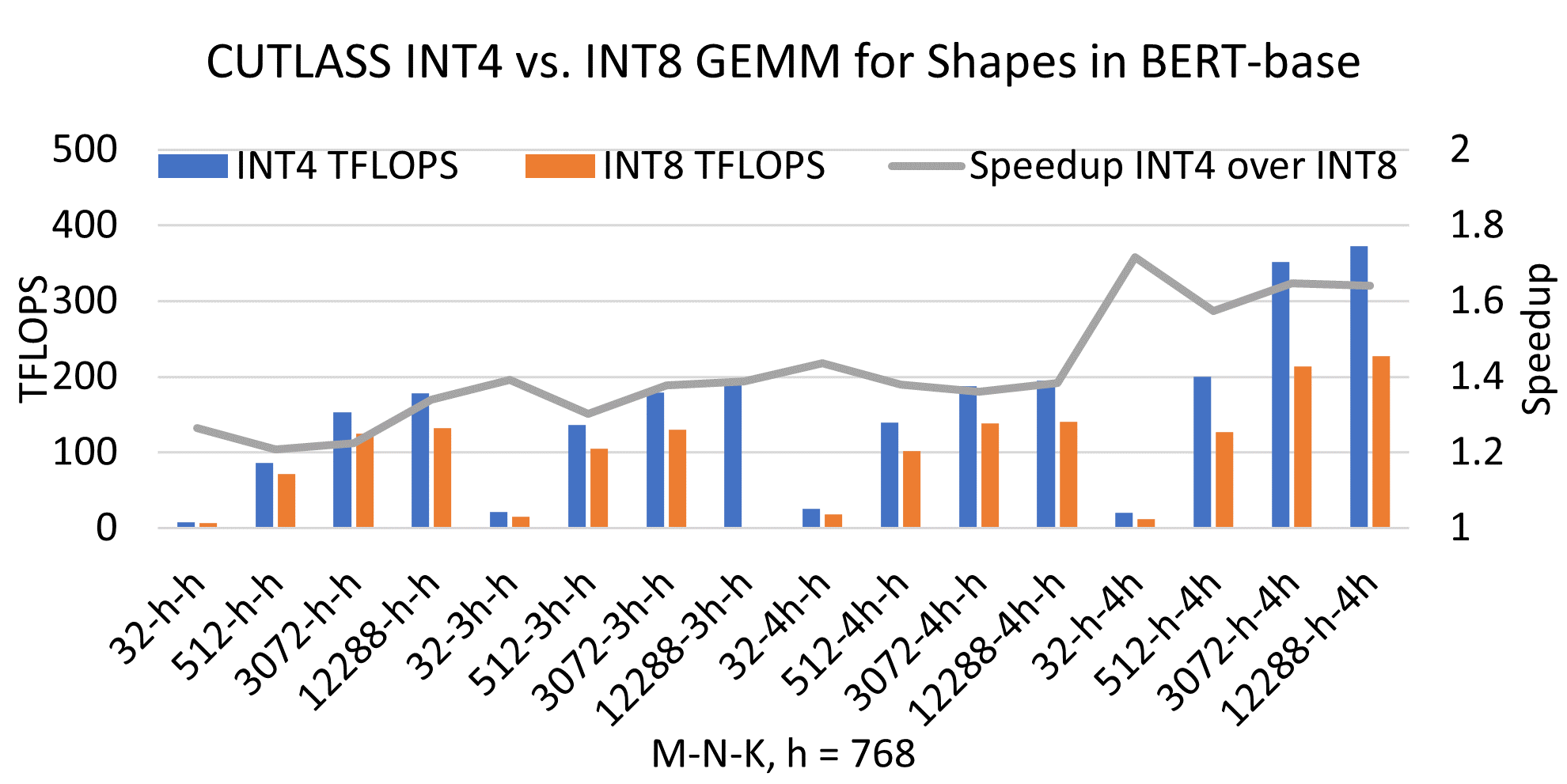}}
\hspace{-.6cm}
\end{subfigure}
\begin{subfigure}{.5\textwidth}
{ \includegraphics[width=1\textwidth]{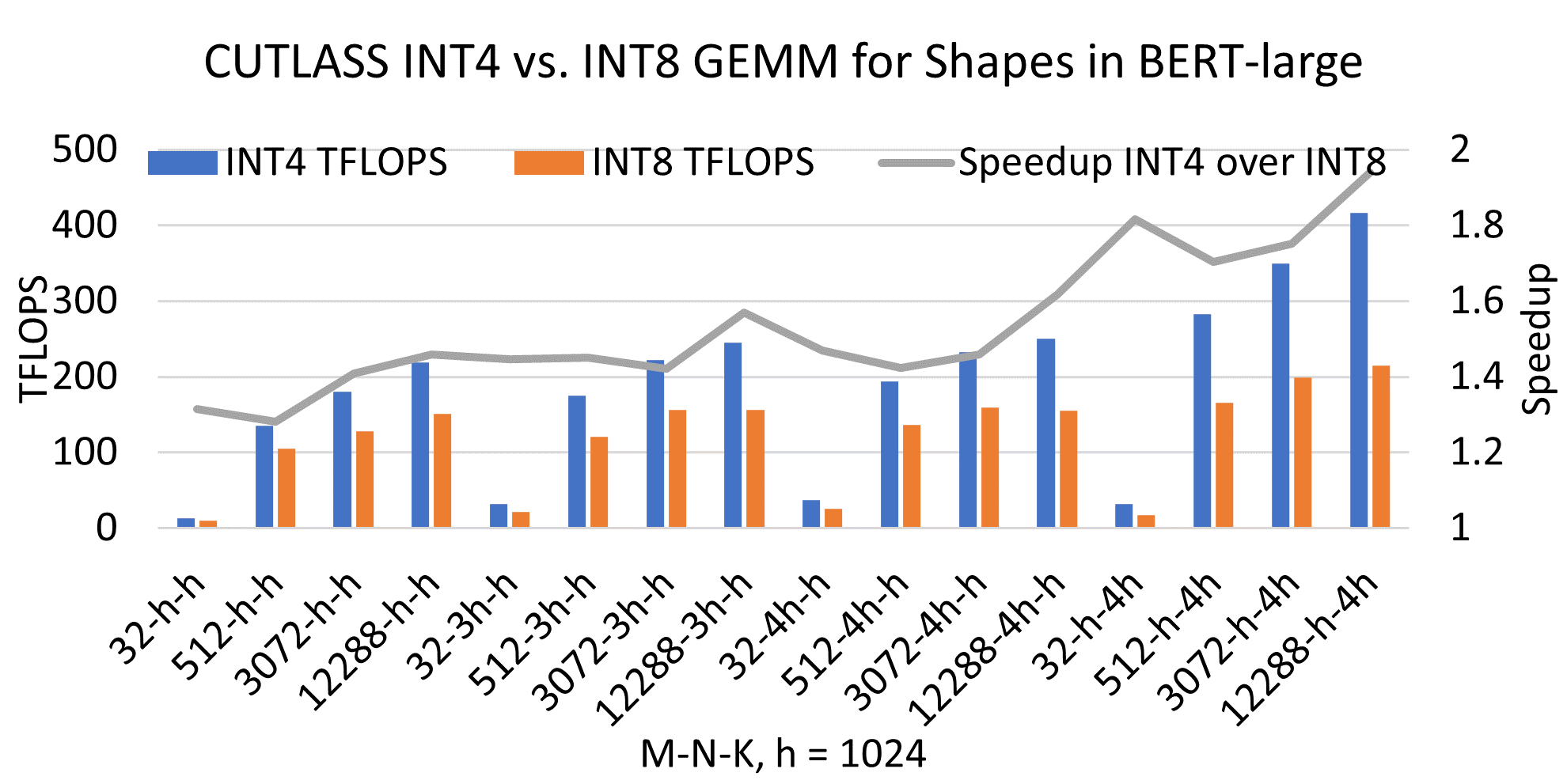}}
\end{subfigure}
 \caption{CUTLASS INT4 vs. INT8 GEMM performance comparison across different batch size$\times$sequence length (M) for BERT-base and BERT-large GEMM shapes (N and K). We use the best GEMM schedule for different inputs identified with the CUTLASS profiler. Left axis shows the throughput achieved (Peak INT8 and INT4 Tensor TOPS is 309.7 and 619.3 TFLOPS on A6000 GPU) and the right axis shows the speedup of INT4 over INT8.}
 \label{figure:gemm_perf}
\end{figure*}

\subsection{INT4 GEMM}\label{sec:int4_gemm}

INT4 Tensor Core performance (peak TFLOPS) theoretically doubles INT8 throughput on supported NVIDIA GPUs. 
However, to achieve the $2\times$ speedup, the GEMM input shapes have to be large enough (being compute-intensive). 
The linear layers that are quantized and computed with INT4 data in the encoder model inference are QKV projection, attention output, MLP intermediate, and MLP output GEMM. 
The GEMM shapes (M-N-K) for these layers are ($bs\times~seq-3h-h$), ($bs\times~seq-h-h$), ($bs\times~seq-4h-h$) and ($bs\times~seq-h-4h$) respectively, where $bs$ and $seq$ are input batch size and sequence length, and $h$ is the model hidden dimension. 
These shapes set the upper-bound performance improvement we can achieve with INT4 over INT8 GEMM for a given model.

 \fref{figure:gemm_perf} shows the performance comparison between INT4 and INT8 GEMM for common shapes in BERT-base and BERT-large model. 
 We can see that the larger the input shape, the higher the speedup. 
 While the INT4 GEMM speedup for BERT-large are overall higher than BERT-base as the model hidden dimension is larger ($1024$ vs. $768$), within a model the four GEMM can have very different achieved INT4 speedup given the same input, i.e., $bs\times~seq$.
 For example, with $bs\times~seq = 12288$ for BERT-large, the attention output GEMM (12288-h-h) only achieves $1.46\times$ speedup while the MLP output GEMM (12288-h-4h) achieves $1.96\times$ when using INT4 over INT8 computation.
 Combining with the quantization/dequantization overhead (see~\sref{sec:e2e_design}), this difference suggests the need for tunable quantization strategies (enable/disable quantization on certain GEMM parts) depending on the input shape.  

\subsection{Holistic Optimizations of End-to-end Inference}\label{sec:e2e_design}

\begin{figure*}[ht]
\begin{subfigure}{0.5\textwidth} {\includegraphics[width=1\linewidth]{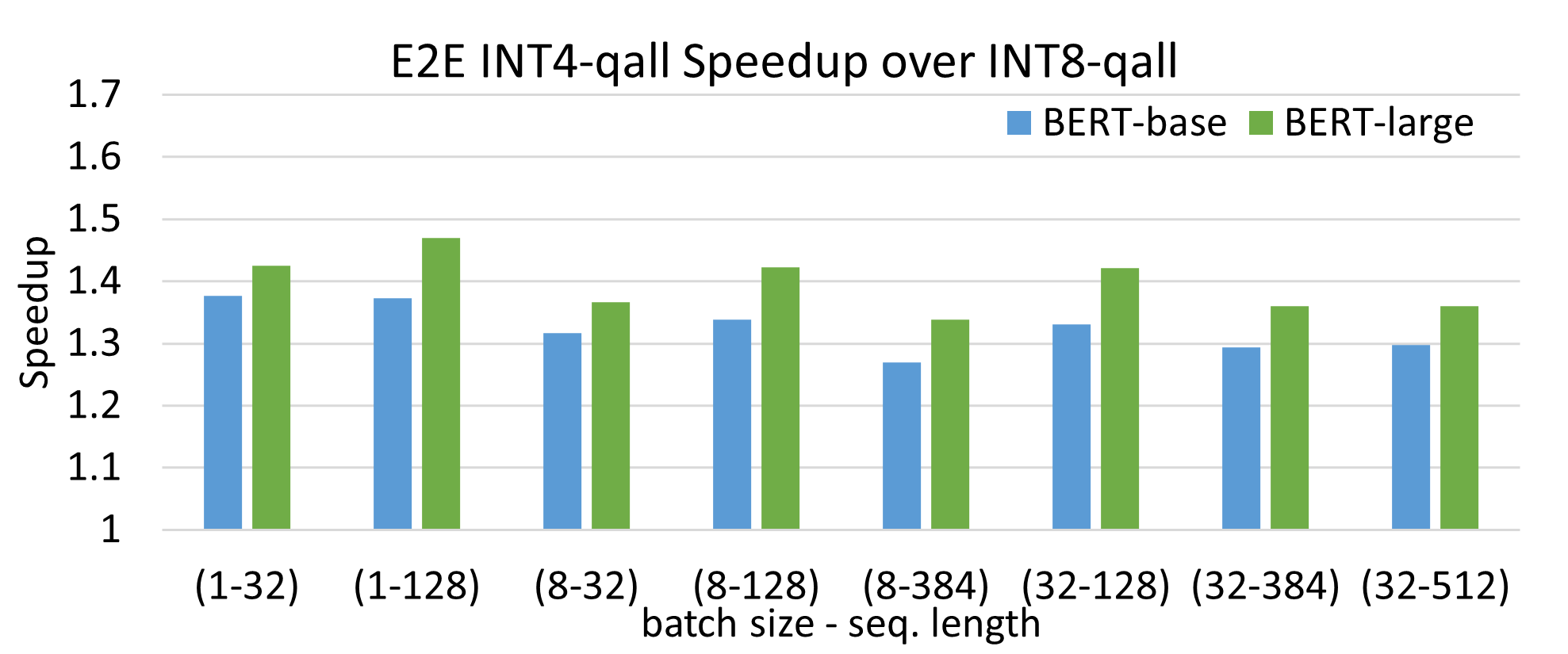}}
\caption{}\label{figure:e2e_i4_i8}
\end{subfigure}
\begin{subfigure}{.5\textwidth}
{ \includegraphics[width=1\textwidth]{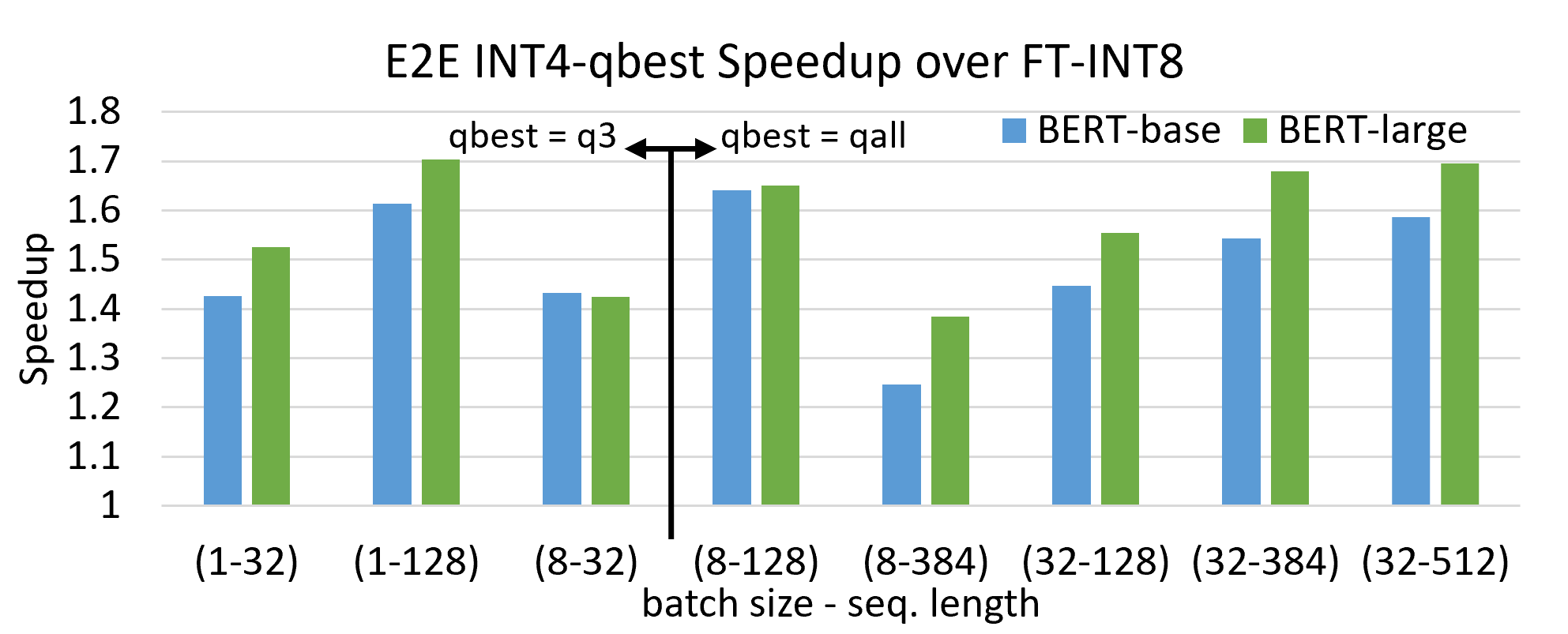}}
\caption{}\label{figure:e2e_i4_fti8}
\end{subfigure}
\caption{E2E latency speedup of (a) our INT4 over INT8 with all four parts quantized (i4-qall and i8-qall), and (b) our INT4 with best quantization strategy (i4-qbest) over Fastertransformer INT8 (FT-i8) on A6000.} 
\label{figure:e2e_perf}
\end{figure*}

\begin{figure*}[ht]
\begin{subfigure}{.5\textwidth} {\includegraphics[width=1\linewidth]{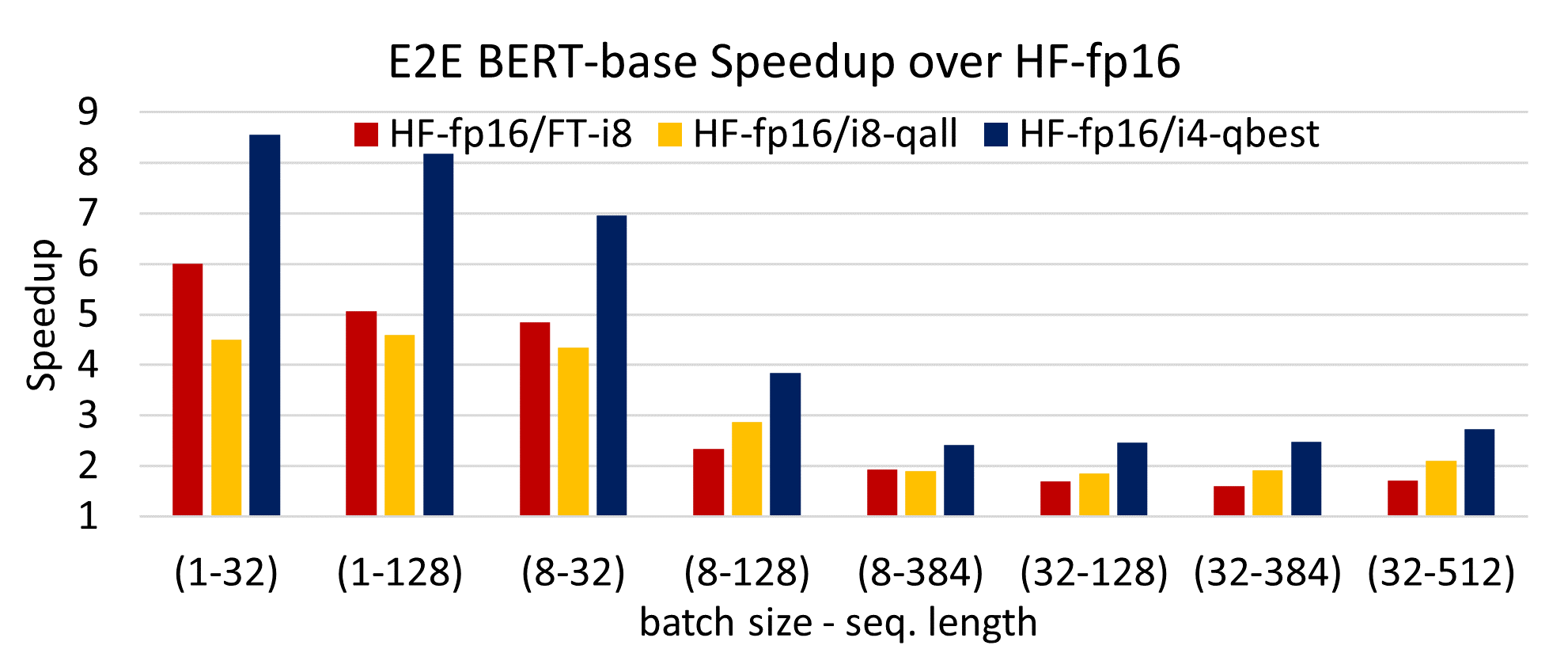}}
\end{subfigure}
\begin{subfigure}{.5\textwidth}
{ \includegraphics[width=1\textwidth]{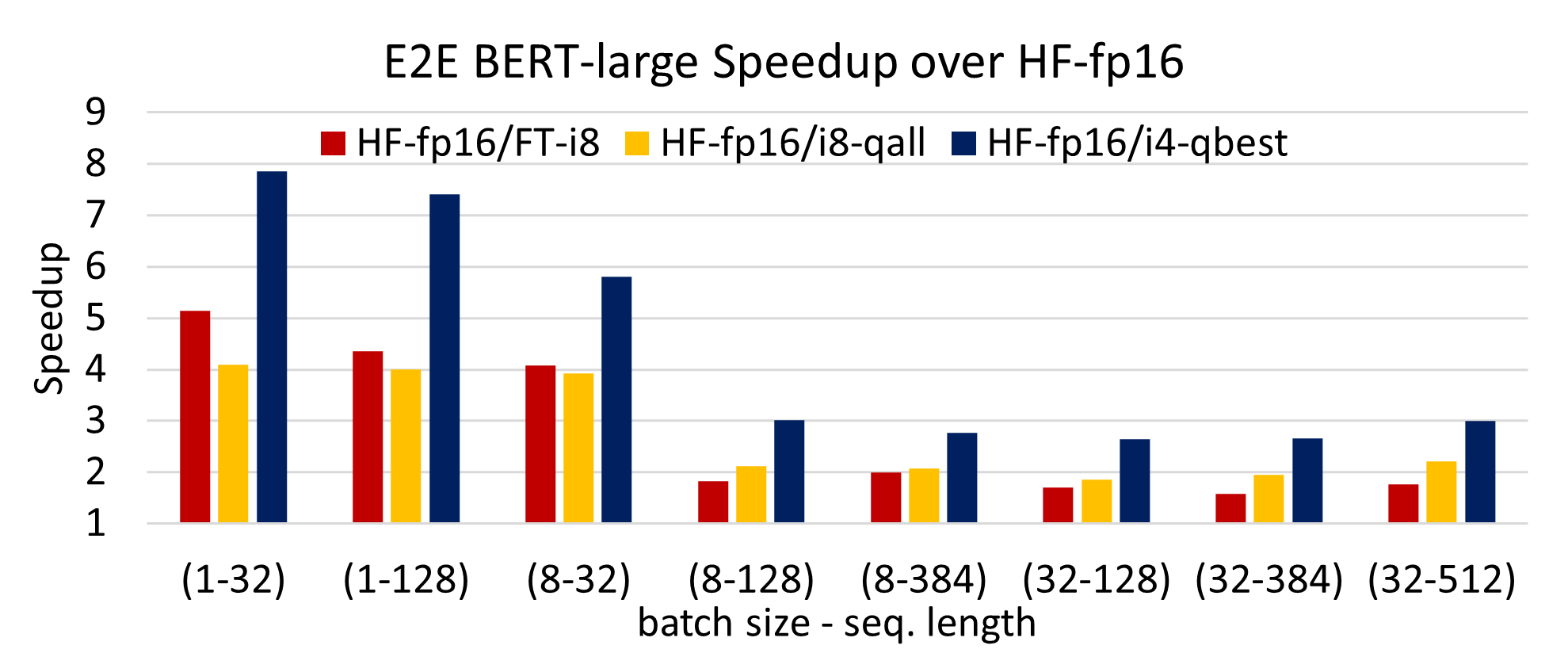}}
\end{subfigure}
\caption{E2E latency speedup of FasterTransformer INT8 (FT-i8), our IN8 with all quantization (q=i8-qall), and our INT4 with best quantization strategy (i4-qbest) over HuggingFace FP16 (HF-fp16) inference.}
 \label{figure:e2e_perf_FP16}
\end{figure*}

While INT4 computation introduces performance improvement for the linear layers, there are other major components in between using FP16 data types (e.g., layer normalization, elementwise operations, etc.). 
The E2E inference requires quantizing/dequantizating the activations before/after the lower-bit GEMM operations. 
Moreover, the improvement from INT4 and the quantization/dequantization overhead are both model- and input-dependent. 
Depending on the deployment scenarios (latency- or throughput-oriented), the optimal quantization strategies can be different.
Thus, maximizing the gain from using INT4 computation requires holistic optimizations of the E2E model inference. 
 
The quantization/dequantization of activations are memory-bound operations and introduce nontrivial overhead.
Similar to ~\citet{yao2022zeroquant}, we fuse the quantization operation for activation with its previous element-bias-add, GELU, or layer normalization operation into a single GPU kernel; 
and fuse the dequantization operation with the INT4 GEMM kernel to avoid extra data movement to global GPU memory.
Since the current PyTorch does not support the INT4 tensor data type yet, we pack INT4 data into INT8 tensors when invoking our customized kernels. 

FlashAttention~\citep{dao2022flashattention} has been shown to largely improve the attention computation performance, especially for large batch sizes and sequence lengths. 
We integrate FlashAttention into our inference pipeline to speed up the attention computation (in FP16).
CUDA graph~\citep{cuda-graph} was introduced by NVIDIA to reduce GPU kernel launching overhead.
For small batch sizes and short sequence lengths, the kernel launching overhead is non-negligible, thus we enable CUDA graph in our inference pipeline to minimize such overhead.

A model deployment scenario can be either latency-sensitive or throughput-oriented, thus different batch sizes and sequence lengths are used for different cases.  
As shown in~\sref{sec:int4_gemm}, the gain from INT4 is input (decides GEMM shapes) dependent. 
The memory-bound quantization/dequantization operations introduce input-dependent (i.e., the size of activations) overhead as well.
Due to the various model sizes (particularly the hidden dimension, h), input shapes, and hardwares, the four linear layers for quantization have different trade-offs between the gain and overhead.
For example, for low $bs\times~seq$ inference with BERT models, quantization of QKV projection, attention output, and MLP output might not result in E2E performance improvement. 
If so, we can skip the quantization of these three parts in inference
(note that using a higher-bit computation data type for a QAT model does not degrade the inference accuracy).

As such, we develop the four model parts as modular components where quantization can be enabled or disabled separately in the inference pipeline. 
Different quantization strategies can be applied given a target scenario and hardware. 
Also, the GEMM schedules used in inference are pre-tuned (with CUTLASS profiler) for the best performance in the deployment environment as well. 

\subsection{End-to-end Inference Performance Results}
We measure the E2E BERT model INT4 (prefixed with \textit{i4-}) and INT8 (prefixed with \textit{i8-}) latency with our inference pipeline and compare it with the HuggingFace FP16 implementation (noted as \textit{HF-fp16}) as well as the SOTA INT8 implementation (noted as \textit{FT-i8}) from NVIDIA FasterTransformer~\citep{fastertransformer}. 
The input batch size and sequence length are selected to cover both latency- and throughput-oriented scenarios. 
We explore different quantization strategies (suffix in name to note what is quantized) with the inference pipeline and show the effectiveness of such tuning.
We use symmetric quantization for the BERT models in the experiments as the earlier section shows no accuracy drop and it is faster than asymmetric quantization because of less required computation for bias term.

\fref{figure:e2e_i4_i8} shows the E2E speedup of our INT4 over our INT8 inference when quantizing all four parts. 
Cross-comparing it with~\fref{figure:gemm_perf} which indicates the upper bound of the E2E INT4-vs-INT8 speedup, we can see that the inference pipeline design achieves well the goal of maximizing the performance gain from using INT4 computation.
\fref{figure:e2e_i4_fti8} compares our best INT4 inference with the Fastertansformer INT8 (using mode-1 as model-2 trades off accuracy for better latency) inference. 
Note that other than the four parts we quantize in our pipeline, Fastertansformer INT8 also quantizes attention computation while we use FP16 FlashAttention (see \sref{sec:e2e_design}).
As annotated, the best quantization strategy for ($bs-seq$) (1-32), (1-128) and (8-32) is to only quantize the MLP intermediate GEMM (q3). 
For larger batch sizes and sequence lengths, the best configuration is to quantize all four parts.
We show that our highly-optimized INT4 inference improves the SOTA BERT model performance by up to $1.7\times$ as compared to FP-INT8, while model quality maintains.

\fref{figure:e2e_perf_FP16} presents the speedup of our inference and FasterTransformer pipelines over HuggingFace FP16 inference, a common baseline for comparison. 
Our INT4 inference is up to $8.5\times$ faster for latency-oriented scenarios and up to $3\times$ for throughput-oriented scenarios.
Note that we focus on maximizing the performance gain from using INT4 computation in this work, thus orthogonal optimizations from FasterTransformer (e.g., padding removal) or other work are applicable to our INT4 inference design, and can further improve the inference performance.

\section{Failure Cases: Understanding the Quality Degradation of INT4 Decoder Models}\label{sec:results-gpt}
For W4A4 GPT models, we have made heavy efforts to tune and distill but their results are still far away from the FP32 counterparts. 
In this section, we present several analyses of the causes of such degradation, including

(1) \textbf{Layer Normalization (LN)}. 
The position of LN is different for encoder and decoder models:  LN for BERT and BART happens after each sublayer’s residual connection (``Post-LN")~\citep{vaswani2017attention}, while LN for GPT models operates at the beginning of each sublayer  before adding to the residual values (``Pre-LN") \citep{xiong2020layer}. 
Compared to Pre-LN, Post-LN removes the mean and variance shift caused by the residual connection and activation functions, which might make the network more robust.
A possible conjecture is that the good quality of INT4 BERT/BART is due to the effect of Post-LN, which thus leads the models to less sensitivity to quantization.  

(2) \textbf{Pretraining Effect}. 
The activation range for decoder models can vary significantly for different layers and for different linear modules. 
A possible conjecture that pretraining with a dataset
of a large scale, such as billions of examples, may exacerbate this issue by introducing more diversity in the input
activations, which could lead to less optimal quantization
performance.


(3) \textbf{Attention Mechanism}. 
GPT models use casual-self-attention mechanism to weight the importance of each word in the input and generate tokens in a sequential manner (autoregressive generation), while BART uses encoder-decoder attention mechanism plus casual-self-attention. 
As such, for the first few generated tokens, BART can still gather information from the encoder-decoder attention which can potentially reduce the quantization error by averaging attention information, while GPT does not have this ability.
\begin{figure}[t]
\centering
\includegraphics[width=0.48\linewidth]{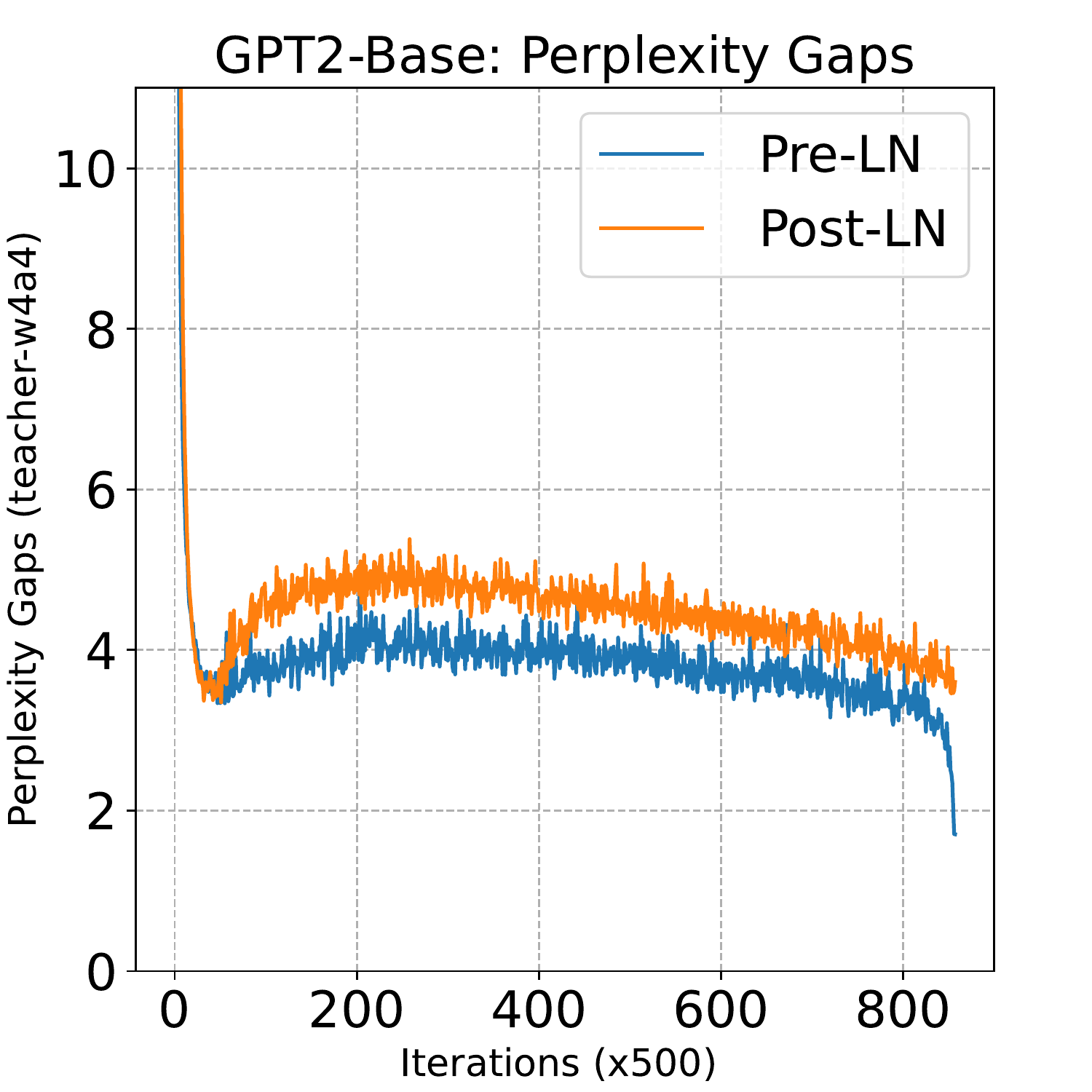}
\includegraphics[width=0.48\linewidth]{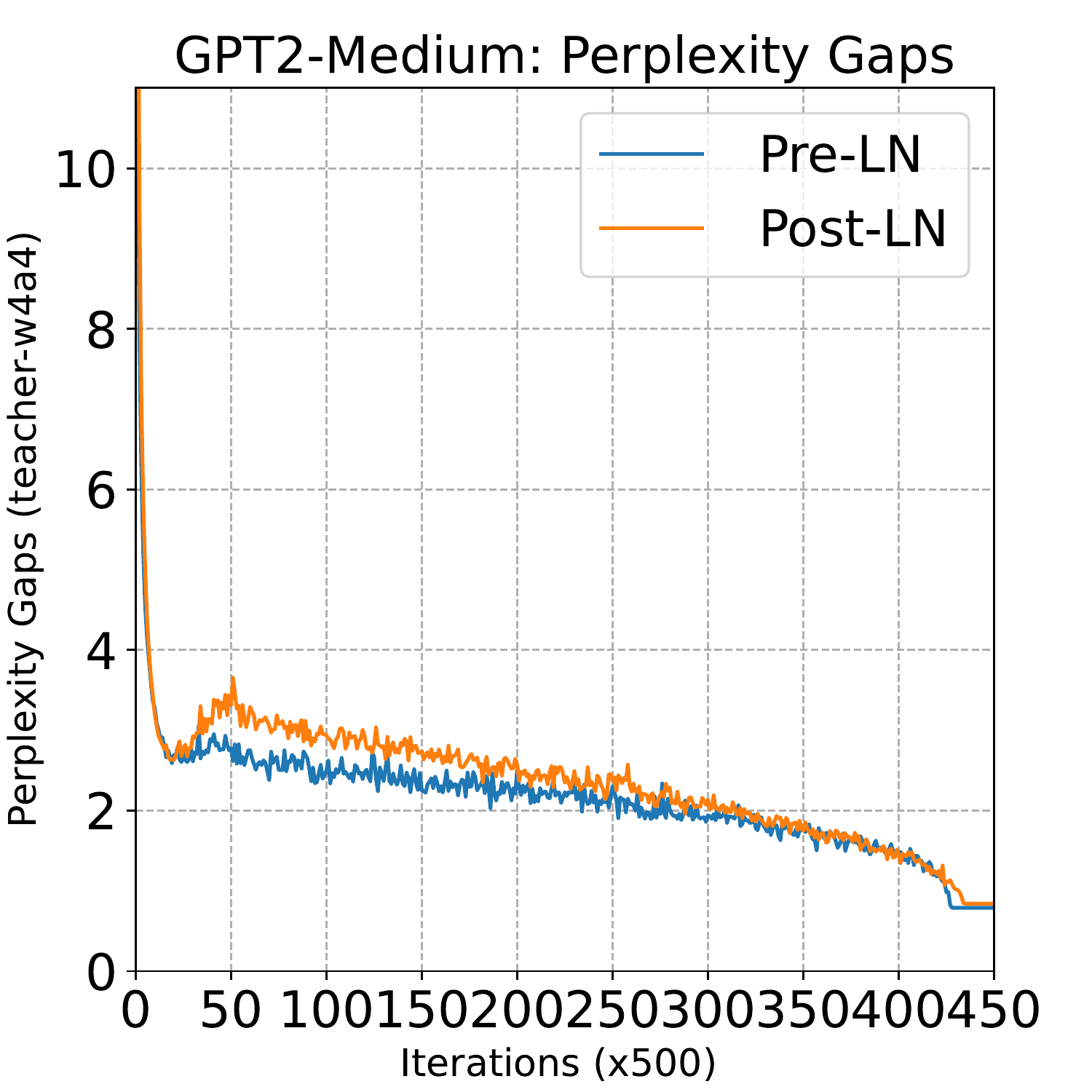}
\caption{The quality gaps between W4A4 and FP32 models, respectively for GPT2-PreLN (blue) and GPT2-PostLN (orange).}\label{figure:gap-ln}
\vspace{-0.4cm}
\end{figure}

\paragraph{Layer Normalization.}  To understand if pre-LN and post-LN lead to a significant difference on the quantization, we design the following experiments:\\
(1) As GPT2 is by default using Pre-LN (GPT2-PreLN),  we construct a model (GPT2-PostLN) by replacing the pre-LN with post-LN. 
In order to have a fair comparison between the quantization results of GPT2-PreLN and GPT2-PostLN, we directly fine-tune both models on Wikitext-103 from scratch, and the perplexities are 17.88 (PreLN) and 18.95 (PostLN) for GPT2-Medium, and 18.76 (PreLN) and 19.46 (PostLN) for GPT2-base.\footnote{Compared to Wikitext-2 and PTB, Wikitext-103 is a considerable larger dataset and thus arrived at a low perplexity even from scratch, closer to results of the pretrained ones.} \\
(2) We take the above FP32 checkpoints and apply QAT with KD to obtain the best W4A4 models. The the perplexities for W4A4 are  18.66 (PreLN) and 19.79 (PostLN) for GPT2-Medium, and 20.46 (PreLN) and 21.73 (PostLN) for GPT2-base.
We then calculate the perplexity gaps between the W4A4 and FP32 models.

 
We report the results in~\fref{figure:gap-ln} of the two perplexity-gap curves for W4A4 and FP32 models, depicted by the blue curve of GPT2-PreLN and orange curve of GPT2-PostLN.
The overlap phenomenon at the end of the training, respectively, demonstrates that LN may not directly affect the performance degradation for decoder-only models..

 \noindent\paragraph{Pretraining Effect.} Despite obtaining negative results on the position of layer normalization, we have identified an intriguing observation in regard to models trained from scratch. 
 Our experiments reveal that the gap between the student and teacher models in terms of perplexity (PPL) is  smaller when training from scratch (20.46 ppl and 18.76 ppl for INT4 and FP32, respectively) as compared to utilizing a pretrained GPT2 model (21.54 ppl for INT4  and 17.46 for FP32). 
 This observation raises questions about the potential negative effort of pretraining in the context of quantization, as the model trained from scratch appears to perform better. 
 \begin{figure}[t]
\centering
\includegraphics[width=0.5\textwidth]{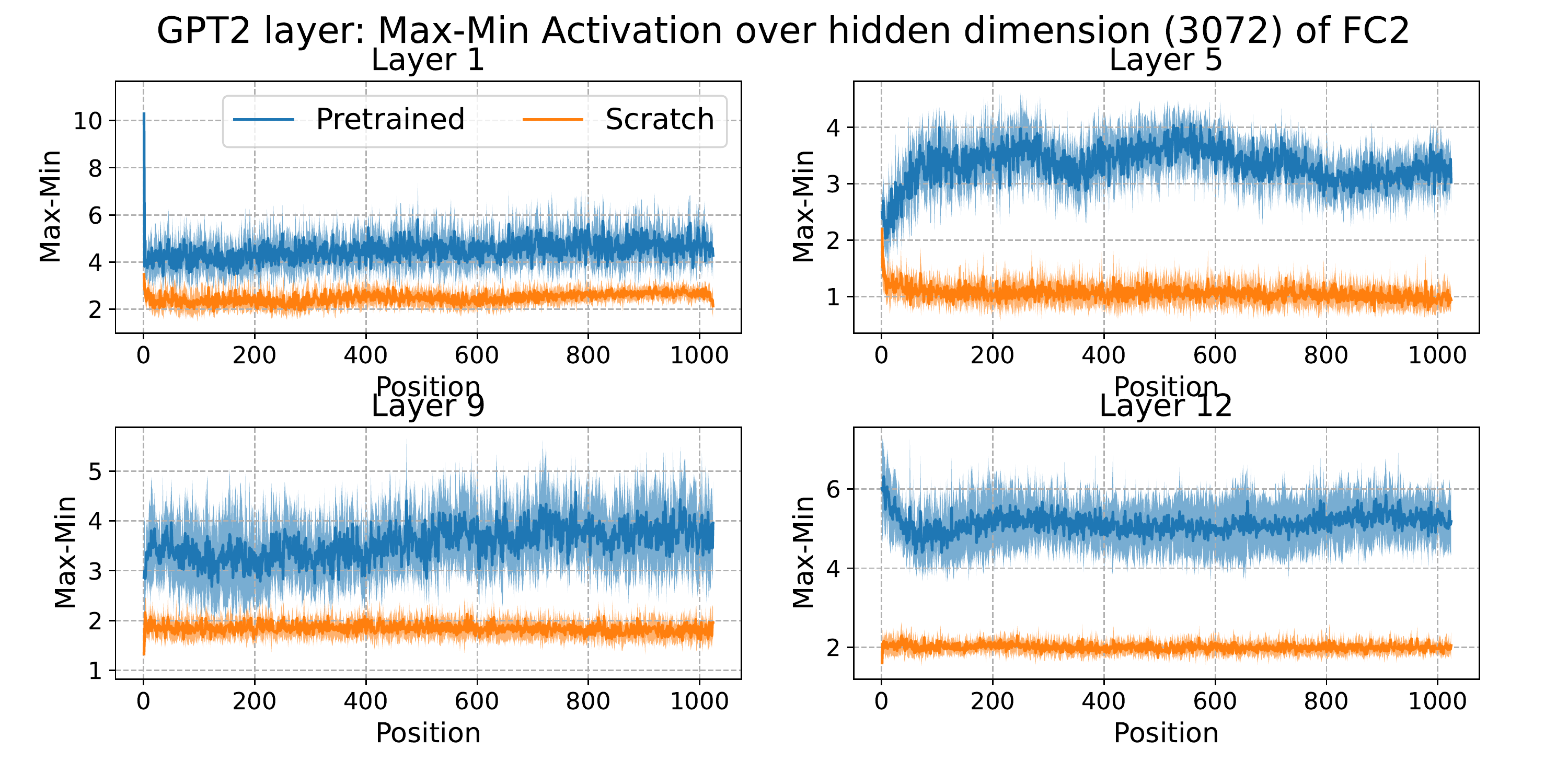} 
\caption{The gaps between the minimum and maximum activations at certain layers (Layer 1, 5, 9, and 12) in the second fully-connected linear module. The gaps are  plotted with respect to position and the average is being taken over 8 batch sizes, with one standard deviation shaded region. }\label{figure:gap-activation}
\end{figure}
To understand this, we compare the position-wise activation range between the fined-tuned models from pretrained checkpoint and from scratch (referred to as ``positional activation").
This provides a token-level understanding on the quantization range. 
The results are shown in~\fref{figure:gap-activation} and
it reveals the higher positional-activation range of the pretrained model as compared to the scratch-trained model.
This further supports the hypothesis that pretraining on large diverse datasets may lead to a wider range of activation values, and thus may be suboptimal for quantization as compared to models trained from scratch.
 
 
\noindent\paragraph{Attention Mechanism.}

To gain insight into the impact of different attention mechanisms (encoder-decoder attention and causal-self-attention) on quantization errors, we conduct a comparison of BART-large and GPT2-medium models. 
We evaluate the ``positional perplexity" of both FP32 and W4A4 models on the CNNDailyMail dataset for BART and Wikitext-2 dataset for GPT. 
The results are depicted in~\fref{figure:gap-ln2}.
We make the following observations: 
\begin{figure}[H]
\centering
\includegraphics[width=0.5\textwidth]{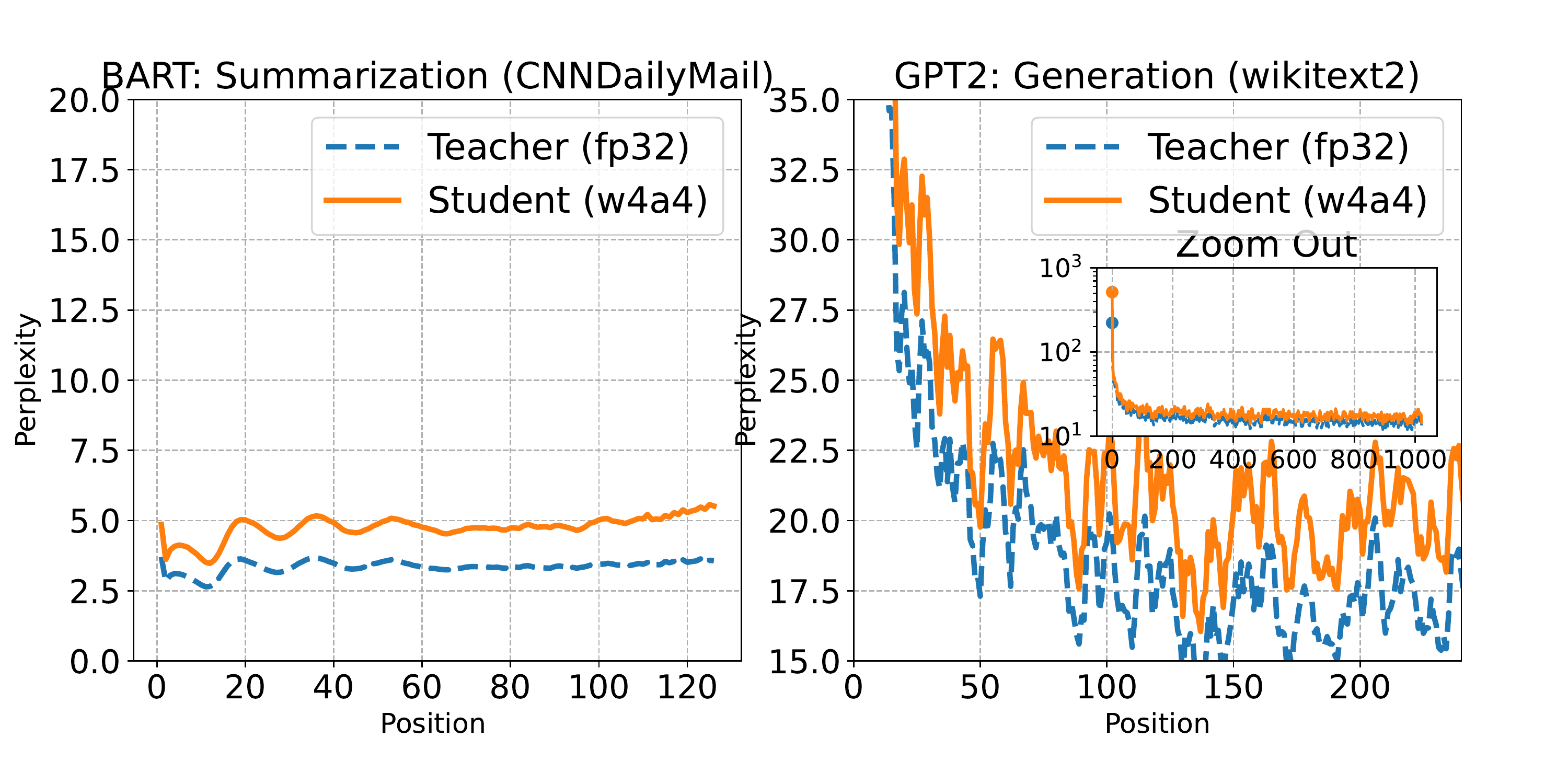}
\caption{The positional perplexity across the full sequence for BART  and GPT2 models. }\label{figure:gap-ln2}
\end{figure}


(1) The curves for GPT, whether it is the teacher or student model, tends to exhibit a downward trend. 
The token losses at early positions are significantly higher than those at later positions. 
Conversely, the curves for both the teacher and student models of BART exhibit a mild upward trend, with token losses at later positions being no better than those at earlier positions. 

(2) The perplexity degradation from quantization for the BART model is small, with a maximum gap of 2.5 ppl at the end of the sequence. 
In contrast, the GPT model experiences large accuracy loss from quantization, with a maximum gap of over 100 ppl at the first tens of tokens of the sequence and around 2 ppl gap at later.

Both phenomena highlight the importance of the additional encoder-decoder attention mechanism. 
For causal-self-attention-only models (i.e., GPT), the next-generated token can only use the information from previous word.
As such, (1) the earlier positions have less information to retrieve, which leads to larger ppl scores; (2) the INT4 model has significant perplexity degradation at the beginning positions compared to FP32 model due to the information noise from quantization.  
Thanks to the encoder-decoder attention, INT4 BART model has relatively (1) stable perplexity for all positions and (2) consistent the positional perplexity degradation as compared to  FP32 counterpart. 

\section{Composability of INT4  Quantization}
\label{sec:discussion}

In this section, we examine the composability of W4A4 to identify techniques that can be used to further accelerate INT4 inference. 
Specifically, we investigate the potential of combining INT4 quantization with other compression techniques, such as pruning and layer reduction. 
Our study is based on the observation that encoder-related models, such as BERT and BART, demonstrate robustness to W4A4 compression as shown in~\tref{table:main-results}. 

\subsection{Composing Semi-structured Pruning with INT4 }\label{sec:results-bert}
We focus on combining semi-structured pruning with W4A4. 
Specifically, we investigate the semi-structured sparsity called Pair-(N:M) which allows for accelerated execution on NVIDIA Ampere GPUs~\citep{mishra2021accelerating,holmes2022compressing}. 
Pair-(N:M) sparsity structure means that there are N zero-entries for every M elements. 
We take BERT-base as an example, as Quantization-Aware Training with Knowledge Distillation for W4A4 models has been shown to lead to better accuracy than its FP32 counterpart. 
We follow the training recipe described in~\cite{wu2022extreme}.



\begin{figure*}[ht]
    \centering
    \includegraphics[width=1.0\textwidth]{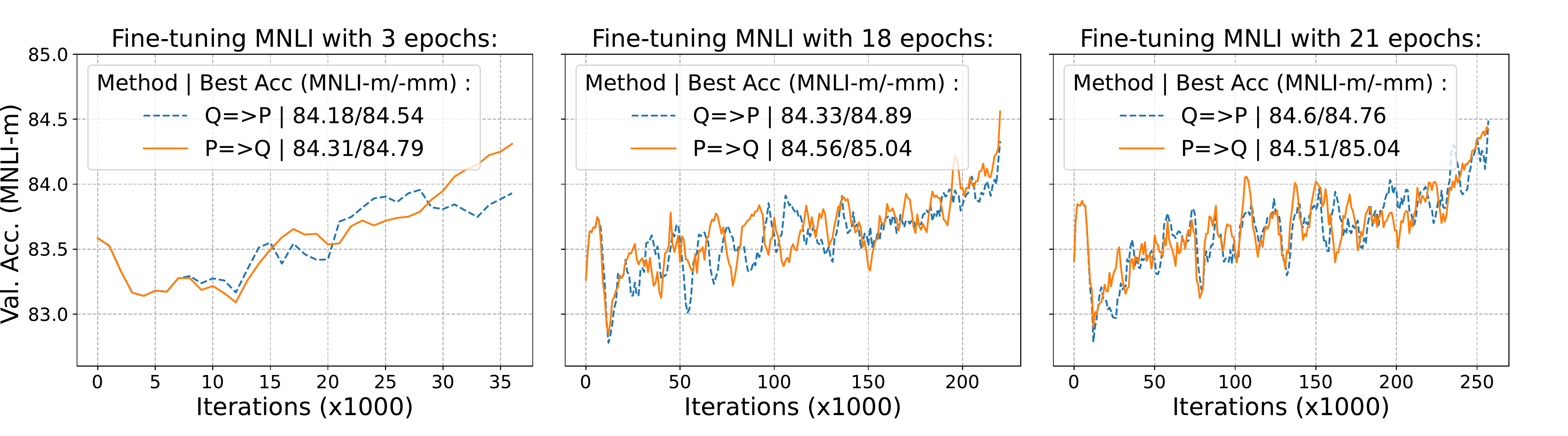}
\caption{The validation accuracy (Val. Acc.) of the W4A4+50\% sparsity (i.e., Pair-(2:4)) BERT-base. We compare the order of pruning and quantization.  Q$=>$P (orange solid curve) means the quantization algorithm is in front of the pruning algorithm. P$=>$Q (blue dash curve) is the opposite.  From left to right plots, the difference is the training epochs (see title). }\label{fig:order}
\end{figure*}
\textbf{Algorithm Design Order between Pruning and INT4.}
When combining the two compression techniques pruning and quantization, a natural question would be the ordering in the forward pass of the two: should we put quantization in front of pruning (e.g. Quant(Prune($\mathbf{W}$) or P$=>$Q), or vice versa (e.g. Prune(Quant($\mathbf{W}$) or Q$=>$P). 
To understand this, we fine-tune on MNLI with different training epochs using simplest  $\ell_1$  pruning method~\citep{han2015learning,han2015deep}. 
$\ell_1$ pruning method prunes those small absolute value to be zero while keeping those large weight value untouched. 
The $\ell_1$ pruning mask is determined by the absolute value of the weight matrix of the teacher models and it remains fixed throughout the training. 

We plot the accuracy on the validation dataset in~\fref{fig:order}. 
As can be seen, for shorter training time,  P$=>$Q is better that Q$=>$P. 
However, the benefits of P$=>$Q start to diminish as we increase the training epochs.
Overall, it is generally recommended to perform pruning before quantization, because pruning removes unnecessary weights from the model.
As such, it can help mitigate the loss of precision caused by quantization and make the quantization process more effective.

With the decision to use the pruning-quantization order, we trained an INT4 BERT-base model with both 50\% and 75\% sparsity and reported the best validation results in~\tref{table:mnli}. 
We found that a 75\% sparsity level results in an accuracy drop of 0.79/1.6 for the MNLI-m/mm tasks. 
Therefore, if maintaining high accuracy is a priority, using a 50\% sparsity level for W4A4 models is recommended. 
In the appendix, we also present the results of applying 50\% sparsity to W4A4 models for 8 GLUE tasks and confirm that the average GLUE scores are similar to those of the original FP32 models.

%
\begin{table}[ht]
\begin{minipage}{.5\textwidth}
\begin{center}
\caption{Quantization(Q)\&Pruning(P) 50\% or 75\% sparsity.} \label{table:mnli}
\vspace{-0.3cm}
\begin{adjustbox}{width=0.99\linewidth}
\begin{tabular}{l|cc|cc}
\toprule
 Tasks          &  Teacher          & Epoch-3 & \multicolumn{2}{c}{Epoch-21: P+Q}          \\
MNLI-& FP32  &  Q only &  50\% sparisty &  75\% sparisty \\\midrule
m/mm                & 84.9/85.6      & 84.8/85.2         & 84.56/85.04      & 84.11/83.99                      \\ \bottomrule                               
\end{tabular}

\end{adjustbox}

\end{center}

 \end{minipage}  
 \end{table}

\subsection{Composing Layer-reduction with INT4}\label{sec:results-bart}
Reducing the depth of a model, also known as layer-reduction, is a more straightforward method to improve inference latency as it requires no modifications to the single-layer implementation (e.g. GPU kernels). 
However, it should be noted that these layer-reduced models may not be able to capture the same level of complexity or learn the same representations as the original models. 
To understand the compatibility of layer-reduction and quantization as well as the trade-off between model depth and quality, we perform detailed study on an encoder-decoder model.

Our implementation of layer-reduction strategies and fine-tuning recipes follows the work in \cite{li2022dq}\footnote{https://github.com/amazon-science/dq-bart}. 
However, there are two key differences: (1) Our quantization algorithm, described in~\sref{sec:methodology}, differs from the one used in \cite{li2022dq}.\footnote{ We did a comparison of the two quantization algorithms and found that the algorithm for INT4 presented in~\sref{sec:methodology} has better accuracy than that in~\cite{li2022dq}} (2) While \cite{li2022dq} uses 8-bit activations, we trained our model with 4-bit activations.

\textbf{More Encoder or More Decoder?} When applying layer-reduction for encoder-decoder model, we need to decide the number of encoder and decoder layers. 
For example, when the depth is fixed at four layers, should we have more encoder layers (3-encoder and 1-decoder), more decoder layers (1-encoder and 3-decoder), or an equal number of layers for both (2-encoder and 2-decoder)? 
We investigate different scenarios of $x$-encoder and $y$-decoder layers, where $x+y\in \{9,7\}$ and $x\in \{6,5,4,3\}$ for the case of $x+y=9$, and $x\in\{3,2,1\}$ for the case of $x+y=4$. We train our models for 10 epochs with a fixed random seed and a learning rate of 5e-5. 

The results are reported in \tref{table:main-bart}. 
A comparison of the results within the same depth (i.e., 9, 7, and 4) reveals that it is beneficial to have more encoder layers than decoder layers, and that the decoder layers should be greater than one. 
Particularly, our experiments demonstrate that the performance of a 9-layer W4A4 BART model (with 6 encoder layers and 3 decoder layers) can be maintained at an acceptable level, which is only 0.6 lower than the 12-layer INT4 on the CNNDailyMail dataset. 
This represents a potential latency improvement of about 50\% for the decoder part while experiencing a minor accuracy drop.

\begin{table}[ht]
\caption{The W4A4 model with layer-reduction. For references, the original W4A4 BART-base (6-encoder and 6-decoder) on CNNDailyMail and XSUM are respectively 44.83/21.67/42.08 and 41.53/18.56/33.62.}\label{table:main-bart}
\vspace{-0.3cm}
\begin{adjustbox}{width=0.999\linewidth}
\centering
\begin{tabular}{l|ccc}
\toprule
Encoder (Decoder)  &  Six (Three)  & Five (Four)	 & Four (Five)	                             \\
CNNDailyMail & \textbf{44.23/21.07/41.58} & 44.15/21.02/41.45 & 43.96/20.9/41.26 \\ 
XSUM         & \textbf{40.61/17.83/32.90}  & 40.30/17.53/32.61  & 40.18/17.47/32.43  \\
\bottomrule
\end{tabular}
\end{adjustbox}
\begin{adjustbox}{width=0.999\linewidth}
\centering
\begin{tabular}{l|cccc|cc}
Encoder (Decoder) & Six (One)  & Five (Two)	 & Four (Three)           \\
CNNDailyMail   & 42.48/19.83/40.13 & \textbf{43.55/20.56/40.99} & 43.48/20.38/40.85\\
XSUM           & 38.23/16.45/31.49 & \textbf{39.52/17.03/31.98} & 39.43/16.89/31.80\\
\bottomrule
\end{tabular}
\end{adjustbox}
\begin{adjustbox}{width=0.999\linewidth}
\centering
\begin{tabular}{l|ccc}
Encoder (Decoder) & Three (One)              &  Two (Two)               &  One (Three)             \\
CNNDailyMail    & 41.26/18.58/38.93 & \textbf{41.83/18.82/39.20}  & 41.40/18.39/38.68  \\
XSUM            & 35.88/14.71/29.34 &  \textbf{36.09/14.39/29.01} & 34.69/13.22/27.64 \\
\bottomrule
\end{tabular}
\end{adjustbox}
\end{table}

\textbf{Summary.} We demonstrate that it is possible to combine INT4 quantization with other compression techniques, like composing INT4 and 50\% Ampere-structure sparisty with aronund 0.5 GLUE points degradation and composing INT4 and 25\% layer reduction without causing much degradation on summarization tasks. 
Fully investing the composability of quantization and other methods is beyond the scope of our paper and we leave it as future work.
\section{Discussion}
\label{sec:conclusions}
\textbf{Conclusion.} Improving the inference efficiency of language models has been increasingly critical given their growing adoption but high compute resource requirements.
While quantization techniques enabling INT8 computation on these language models have been well explored recently, how to fully unlock the INT4 computation power of GPU is an emerging and unanswered question. 
In this work, we thoroughly investigate the feasibility of applying INT4 quantization to language models, and our INT4 encoder inference pipeline shows an up to $1.7\times$ latency improvement over SOTA INT8 inference. 
 We provide an analysis of the accuracy drop for decoder models when using INT4 quantization, and study the composability of W4A4 for encoder-related model with other compression techniques.

\textbf{Limitation.} Our approach build upon existing techniques, including distillation-assisted quantization, thus limiting its novelty.  Although we introduced an analysis of failure cases in decoder models, some of these assessments may not be comprehensive, and drawn conclusions could be subject to non-optimized conditions, such as the optimization of the quantization clipping value. Furthermore, the study lacks an end-to-end performance measure on INT4 decoder models. We argued that implementation should only follow the resolution of accuracy issues, but including a performance measure could provide valuable insights about pursuing INT4 efforts, considering the unique memory-bound challenge posed by decoder models. We acknowledge these limitations in the pursuit of an honest and accurate discourse.

\section*{Acknowledgement}
This research was conducted within the supportive environment of the DeepSpeed team at Microsoft, whose invaluable assistance was instrumental to this project. We are immensely grateful for the insightful feedback from the anonymous reviewers from the International Conference on Machine Learning (ICML) 2023. We would also like to express our appreciation to Dr. Dan Alistarh for highlighting his recent works, which facilitated our research.

\clearpage
{
\bibliographystyle{ICML2023_format/icml2023}
\bibliography{ref.bib}

\begin{thebibliography}{72}
\providecommand{\natexlab}[1]{#1}
\providecommand{\url}[1]{\texttt{#1}}
\expandafter\ifx\csname urlstyle\endcsname\relax
  \providecommand{\doi}[1]{doi: #1}\else
  \providecommand{\doi}{doi: \begingroup \urlstyle{rm}\Url}\fi

\bibitem[Abdolrashidi et~al.(2021)Abdolrashidi, Wang, Agrawal, Malmaud,
  Rybakov, Leichner, and Lew]{abdolrashidi2021pareto}
Abdolrashidi, A., Wang, L., Agrawal, S., Malmaud, J., Rybakov, O., Leichner,
  C., and Lew, L.
\newblock Pareto-optimal quantized resnet is mostly 4-bit.
\newblock In \emph{Proceedings of the IEEE/CVF Conference on Computer Vision
  and Pattern Recognition}, pp.\  3091--3099, 2021.

\bibitem[Aminabadi et~al.(2022)Aminabadi, Rajbhandari, Zhang, Awan, Li, Li,
  Zheng, Rasley, Smith, Ruwase, et~al.]{aminabadi2022deepspeed}
Aminabadi, R.~Y., Rajbhandari, S., Zhang, M., Awan, A.~A., Li, C., Li, D.,
  Zheng, E., Rasley, J., Smith, S., Ruwase, O., et~al.
\newblock Deepspeed inference: Enabling efficient inference of transformer
  models at unprecedented scale.
\newblock \emph{arXiv preprint arXiv:2207.00032}, 2022.

\bibitem[Bai et~al.(2020)Bai, Zhang, Hou, Shang, Jin, Jiang, Liu, Lyu, and
  King]{bai2020binarybert}
Bai, H., Zhang, W., Hou, L., Shang, L., Jin, J., Jiang, X., Liu, Q., Lyu, M.,
  and King, I.
\newblock Binarybert: Pushing the limit of bert quantization.
\newblock \emph{arXiv preprint arXiv:2012.15701}, 2020.

\bibitem[Bengio et~al.(2013)Bengio, L{\'e}onard, and
  Courville]{bengio2013estimating}
Bengio, Y., L{\'e}onard, N., and Courville, A.
\newblock Estimating or propagating gradients through stochastic neurons for
  conditional computation.
\newblock \emph{arXiv preprint arXiv:1308.3432}, 2013.

\bibitem[Cer et~al.(2017)Cer, Diab, Agirre, Lopez-Gazpio, and
  Specia]{cer2017semeval}
Cer, D., Diab, M., Agirre, E., Lopez-Gazpio, I., and Specia, L.
\newblock Semeval-2017 task 1: Semantic textual similarity-multilingual and
  cross-lingual focused evaluation.
\newblock \emph{arXiv preprint arXiv:1708.00055}, 2017.

\bibitem[Chung et~al.(2020)Chung, Kim, Choi, Kwon, Jeon, Park, Kim, and
  Lee]{chung2020extremely}
Chung, I., Kim, B., Choi, Y., Kwon, S.~J., Jeon, Y., Park, B., Kim, S., and
  Lee, D.
\newblock Extremely low bit transformer quantization for on-device neural
  machine translation.
\newblock \emph{arXiv preprint arXiv:2009.07453}, 2020.

\bibitem[Dagan et~al.(2013)Dagan, Roth, Sammons, and
  Zanzotto]{dagan2013recognizing}
Dagan, I., Roth, D., Sammons, M., and Zanzotto, F.~M.
\newblock Recognizing textual entailment: Models and applications.
\newblock \emph{Synthesis Lectures on Human Language Technologies}, 6\penalty0
  (4):\penalty0 1--220, 2013.

\bibitem[Dao et~al.(2022)Dao, Fu, Ermon, Rudra, and
  R{\'e}]{dao2022flashattention}
Dao, T., Fu, D.~Y., Ermon, S., Rudra, A., and R{\'e}, C.
\newblock Flashattention: Fast and memory-efficient exact attention with
  io-awareness.
\newblock \emph{arXiv preprint arXiv:2205.14135}, 2022.

\bibitem[Dehghani et~al.(2018)Dehghani, Gouws, Vinyals, Uszkoreit, and
  Kaiser]{dehghani2018universal}
Dehghani, M., Gouws, S., Vinyals, O., Uszkoreit, J., and Kaiser, {\L}.
\newblock Universal transformers.
\newblock \emph{arXiv preprint arXiv:1807.03819}, 2018.

\bibitem[Dettmers \& Zettlemoyer(2022)Dettmers and
  Zettlemoyer]{dettmers2022case}
Dettmers, T. and Zettlemoyer, L.
\newblock The case for 4-bit precision: k-bit inference scaling laws.
\newblock \emph{arXiv preprint arXiv:2212.09720}, 2022.

\bibitem[Dettmers et~al.(2022{\natexlab{a}})Dettmers, Lewis, Belkada, and
  Zettlemoyer]{dettmers2022gptint}
Dettmers, T., Lewis, M., Belkada, Y., and Zettlemoyer, L.
\newblock {GPT}3.int8(): 8-bit matrix multiplication for transformers at scale.
\newblock In Oh, A.~H., Agarwal, A., Belgrave, D., and Cho, K. (eds.),
  \emph{Advances in Neural Information Processing Systems}, 2022{\natexlab{a}}.
\newblock URL \url{https://openreview.net/forum?id=dXiGWqBoxaD}.

\bibitem[Dettmers et~al.(2022{\natexlab{b}})Dettmers, Lewis, Belkada, and
  Zettlemoyer]{dettmers2022llm}
Dettmers, T., Lewis, M., Belkada, Y., and Zettlemoyer, L.
\newblock Llm. int8 (): 8-bit matrix multiplication for transformers at scale.
\newblock \emph{arXiv preprint arXiv:2208.07339}, 2022{\natexlab{b}}.

\bibitem[Dolan \& Brockett(2005)Dolan and Brockett]{dolan2005automatically}
Dolan, W.~B. and Brockett, C.
\newblock Automatically constructing a corpus of sentential paraphrases.
\newblock In \emph{Proceedings of the Third International Workshop on
  Paraphrasing (IWP2005)}, 2005.

\bibitem[Dong et~al.(2019)Dong, Yao, Gholami, Mahoney, and
  Keutzer]{dong2019hawq}
Dong, Z., Yao, Z., Gholami, A., Mahoney, M.~W., and Keutzer, K.
\newblock {HAWQ}: Hessian aware quantization of neural networks with
  mixed-precision.
\newblock In \emph{Proceedings of the IEEE International Conference on Computer
  Vision}, pp.\  293--302, 2019.

\bibitem[Fan et~al.(2019)Fan, Grave, and Joulin]{fan2019reducing}
Fan, A., Grave, E., and Joulin, A.
\newblock Reducing transformer depth on demand with structured dropout.
\newblock \emph{arXiv preprint arXiv:1909.11556}, 2019.

\bibitem[Frantar \& Alistarh(2022)Frantar and Alistarh]{frantar2022optimal}
Frantar, E. and Alistarh, D.
\newblock Optimal brain compression: A framework for accurate post-training
  quantization and pruning.
\newblock \emph{arXiv preprint arXiv:2208.11580}, 2022.

\bibitem[Frantar et~al.(2022)Frantar, Ashkboos, Hoefler, and
  Alistarh]{frantar2022gptq}
Frantar, E., Ashkboos, S., Hoefler, T., and Alistarh, D.
\newblock Gptq: Accurate post-training quantization for generative pre-trained
  transformers.
\newblock \emph{arXiv preprint arXiv:2210.17323}, 2022.

\bibitem[Gholami et~al.(2021)Gholami, Kim, Dong, Yao, Mahoney, and
  Keutzer]{gholami2021survey}
Gholami, A., Kim, S., Dong, Z., Yao, Z., Mahoney, M.~W., and Keutzer, K.
\newblock A survey of quantization methods for efficient neural network
  inference.
\newblock \emph{arXiv preprint arXiv:2103.13630}, 2021.

\bibitem[Gordon et~al.(2020)Gordon, Duh, and Andrews]{gordon2020compressing}
Gordon, M.~A., Duh, K., and Andrews, N.
\newblock Compressing bert: Studying the effects of weight pruning on transfer
  learning.
\newblock \emph{arXiv preprint arXiv:2002.08307}, 2020.

\bibitem[Han et~al.(2015)Han, Pool, Tran, and Dally]{han2015learning}
Han, S., Pool, J., Tran, J., and Dally, W.
\newblock Learning both weights and connections for efficient neural network.
\newblock In \emph{Advances in neural information processing systems}, pp.\
  1135--1143, 2015.

\bibitem[Han et~al.(2016)Han, Mao, and Dally]{han2015deep}
Han, S., Mao, H., and Dally, W.~J.
\newblock Deep compression: Compressing deep neural networks with pruning,
  trained quantization and huffman coding.
\newblock \emph{International Conference on Learning Representations}, 2016.

\bibitem[Han et~al.(2020)Han, Zhang, Li, Liu, Tian, Xie, and
  Shan]{han2020convolutional}
Han, T., Zhang, T., Li, D., Liu, G., Tian, L., Xie, D., and Shan, Y.~S.
\newblock Convolutional neural network with int4 optimization on xilinx
  devices.
\newblock \emph{Xilinx White Paper, WP521}, 2020.

\bibitem[Hermann et~al.(2015)Hermann, Kocisky, Grefenstette, Espeholt, Kay,
  Suleyman, and Blunsom]{hermann2015teaching}
Hermann, K.~M., Kocisky, T., Grefenstette, E., Espeholt, L., Kay, W., Suleyman,
  M., and Blunsom, P.
\newblock Teaching machines to read and comprehend.
\newblock \emph{arXiv preprint arXiv:1506.03340}, 2015.

\bibitem[Hinton et~al.(2014)Hinton, Vinyals, and Dean]{hinton2015distilling}
Hinton, G., Vinyals, O., and Dean, J.
\newblock Distilling the knowledge in a neural network.
\newblock \emph{Workshop paper in NIPS}, 2014.

\bibitem[Holmes et~al.(2022)Holmes, Zhang, He, and Wu]{holmes2022compressing}
Holmes, C., Zhang, M., He, Y., and Wu, B.
\newblock Compressing pre-trained transformers via low-bit nxm sparsity for
  natural language understanding.
\newblock \emph{arXiv preprint arXiv:2206.15014}, 2022.

\bibitem[Hu et~al.(2021)Hu, Wallis, Allen-Zhu, Li, Wang, Wang, Chen,
  et~al.]{hu2021lora}
Hu, E.~J., Wallis, P., Allen-Zhu, Z., Li, Y., Wang, S., Wang, L., Chen, W.,
  et~al.
\newblock Lora: Low-rank adaptation of large language models.
\newblock In \emph{International Conference on Learning Representations}, 2021.

\bibitem[Iyer et~al.(2017)Iyer, Dandekar, and Csernai]{iyer2017first}
Iyer, S., Dandekar, N., and Csernai, K.
\newblock First quora dataset release: Question pairs.(2017).
\newblock \emph{URL https://data. quora.
  com/First-Quora-Dataset-Release-Question-Pairs}, 2017.

\bibitem[Jiao et~al.(2019)Jiao, Yin, Shang, Jiang, Chen, Li, Wang, and
  Liu]{jiao2019tinybert}
Jiao, X., Yin, Y., Shang, L., Jiang, X., Chen, X., Li, L., Wang, F., and Liu,
  Q.
\newblock Tinybert: Distilling bert for natural language understanding.
\newblock \emph{arXiv preprint arXiv:1909.10351}, 2019.

\bibitem[Kim et~al.(2021)Kim, Gholami, Yao, Mahoney, and Keutzer]{kim2021bert}
Kim, S., Gholami, A., Yao, Z., Mahoney, M.~W., and Keutzer, K.
\newblock I-bert: Integer-only bert quantization.
\newblock In \emph{International conference on machine learning}, pp.\
  5506--5518. PMLR, 2021.

\bibitem[Lagunas et~al.(2021)Lagunas, Charlaix, Sanh, and
  Rush]{lagunas2021block}
Lagunas, F., Charlaix, E., Sanh, V., and Rush, A.~M.
\newblock Block pruning for faster transformers.
\newblock In \emph{Proceedings of the 2021 Conference on Empirical Methods in
  Natural Language Processing}, pp.\  10619--10629, 2021.

\bibitem[Lambda(2023)]{lambda-box}
Lambda.
\newblock {GPU workstation for deep learning}.
\newblock \url{https://lambdalabs.com/gpu-workstations/vector}, January 2023.

\bibitem[Lan et~al.(2019)Lan, Chen, Goodman, Gimpel, Sharma, and
  Soricut]{lan2019albert}
Lan, Z., Chen, M., Goodman, S., Gimpel, K., Sharma, P., and Soricut, R.
\newblock Albert: A lite bert for self-supervised learning of language
  representations.
\newblock \emph{arXiv preprint arXiv:1909.11942}, 2019.

\bibitem[LeCun et~al.(1990)LeCun, Denker, and Solla]{lecun1990optimal}
LeCun, Y., Denker, J.~S., and Solla, S.~A.
\newblock Optimal brain damage.
\newblock In \emph{Advances in neural information processing systems}, pp.\
  598--605, 1990.

\bibitem[Lewis et~al.(2020)Lewis, Liu, Goyal, Ghazvininejad, Mohamed, Levy,
  Stoyanov, and Zettlemoyer]{lewis2020bart}
Lewis, M., Liu, Y., Goyal, N., Ghazvininejad, M., Mohamed, A., Levy, O.,
  Stoyanov, V., and Zettlemoyer, L.
\newblock Bart: Denoising sequence-to-sequence pre-training for natural
  language generation, translation, and comprehension.
\newblock In \emph{Proceedings of the 58th Annual Meeting of the Association
  for Computational Linguistics}, pp.\  7871--7880, 2020.

\bibitem[Li et~al.(2016{\natexlab{a}})Li, Zhang, and Liu]{li2016ternary}
Li, F., Zhang, B., and Liu, B.
\newblock Ternary weight networks.
\newblock \emph{arXiv preprint arXiv:1605.04711}, 2016{\natexlab{a}}.

\bibitem[Li et~al.(2016{\natexlab{b}})Li, Kadav, Durdanovic, Samet, and
  Graf]{li2016pruning}
Li, H., Kadav, A., Durdanovic, I., Samet, H., and Graf, H.~P.
\newblock Pruning filters for efficient convnets.
\newblock \emph{arXiv preprint arXiv:1608.08710}, 2016{\natexlab{b}}.

\bibitem[Li et~al.(2022)Li, Wang, Tan, Nallapati, Bhatia, Arnold, Xiang, and
  Roth]{li2022dq}
Li, Z., Wang, Z., Tan, M., Nallapati, R., Bhatia, P., Arnold, A., Xiang, B.,
  and Roth, D.
\newblock Dq-bart: Efficient sequence-to-sequence model via joint distillation
  and quantization.
\newblock In \emph{Proceedings of the 60th Annual Meeting of the Association
  for Computational Linguistics (Volume 2: Short Papers)}, pp.\  203--211,
  2022.

\bibitem[Liu et~al.(2021)Liu, Wang, Han, Zhang, Ma, and Gao]{liu2021post}
Liu, Z., Wang, Y., Han, K., Zhang, W., Ma, S., and Gao, W.
\newblock Post-training quantization for vision transformer.
\newblock \emph{Advances in Neural Information Processing Systems}, 34, 2021.

\bibitem[Mao et~al.(2017)Mao, Han, Pool, Li, Liu, Wang, and
  Dally]{mao2017exploring}
Mao, H., Han, S., Pool, J., Li, W., Liu, X., Wang, Y., and Dally, W.~J.
\newblock Exploring the regularity of sparse structure in convolutional neural
  networks.
\newblock \emph{Workshop paper in CVPR}, 2017.

\bibitem[Mao et~al.(2020)Mao, Wang, Wu, Zhang, Wang, Yang, Zhang, Tong, and
  Bai]{mao2020ladabert}
Mao, Y., Wang, Y., Wu, C., Zhang, C., Wang, Y., Yang, Y., Zhang, Q., Tong, Y.,
  and Bai, J.
\newblock Ladabert: Lightweight adaptation of bert through hybrid model
  compression.
\newblock \emph{arXiv preprint arXiv:2004.04124}, 2020.

\bibitem[Marcinkiewicz(1994)]{marcinkiewicz1994building}
Marcinkiewicz, M.~A.
\newblock Building a large annotated corpus of english: The penn treebank.
\newblock \emph{Using Large Corpora}, pp.\  273, 1994.

\bibitem[Merity et~al.(2017)Merity, Xiong, Bradbury, and
  Socher]{merity2016pointer}
Merity, S., Xiong, C., Bradbury, J., and Socher, R.
\newblock Pointer sentinel mixture models.
\newblock In \emph{International Conference on Learning Representations}, 2017.

\bibitem[Michel et~al.(2019)Michel, Levy, and Neubig]{michel2019sixteen}
Michel, P., Levy, O., and Neubig, G.
\newblock Are sixteen heads really better than one?
\newblock \emph{arXiv preprint arXiv:1905.10650}, 2019.

\bibitem[Micikevicius et~al.(2018)Micikevicius, Narang, Alben, Diamos, Elsen,
  Garcia, Ginsburg, Houston, Kuchaiev, Venkatesh,
  et~al.]{micikevicius2018mixed}
Micikevicius, P., Narang, S., Alben, J., Diamos, G., Elsen, E., Garcia, D.,
  Ginsburg, B., Houston, M., Kuchaiev, O., Venkatesh, G., et~al.
\newblock Mixed precision training.
\newblock In \emph{International Conference on Learning Representations}, 2018.

\bibitem[Mishra et~al.(2021)Mishra, Latorre, Pool, Stosic, Stosic, Venkatesh,
  Yu, and Micikevicius]{mishra2021accelerating}
Mishra, A., Latorre, J.~A., Pool, J., Stosic, D., Stosic, D., Venkatesh, G.,
  Yu, C., and Micikevicius, P.
\newblock Accelerating sparse deep neural networks.
\newblock \emph{arXiv preprint arXiv:2104.08378}, 2021.

\bibitem[Narayan et~al.(2018)Narayan, Martins, Sordoni, Bachman, Courville, and
  Bengio]{narayan2018don}
Narayan, S., Martins, A., Sordoni, A., Bachman, P., Courville, A., and Bengio,
  Y.
\newblock Don't give me the details, just the summary!: topic-aware
  convolutional neural networks for extreme summarization.
\newblock In \emph{Proceedings of the 2018 Conference on Empirical Methods in
  Natural Language Processing}, pp.\  3706--3716, 2018.

\bibitem[NVIDIA(2017)]{cutlass}
NVIDIA.
\newblock {CUTLASS: Fast Linear Algebra in CUDA C++}.
\newblock \url{https://developer.nvidia.com/blog/cutlass-linear-algebra-cuda/},
  December 2017.

\bibitem[NVIDIA(2021)]{cuda-graph}
NVIDIA.
\newblock {Employing CUDA Graphs in a Dynamic Environment}.
\newblock
  \url{https://developer.nvidia.com/blog/employing-cuda-graphs-in-a-dynamic-environment/},
  November 2021.

\bibitem[NVIDIA(2023)]{fastertransformer}
NVIDIA.
\newblock {FasterTransformer}.
\newblock \url{https://github.com/NVIDIA/FasterTransformer}, January 2023.

\bibitem[Polino et~al.(2018)Polino, Pascanu, and Alistarh]{polino2018model}
Polino, A., Pascanu, R., and Alistarh, D.
\newblock Model compression via distillation and quantization.
\newblock \emph{arXiv preprint arXiv:1802.05668}, 2018.

\bibitem[Radford et~al.(2019)Radford, Wu, Child, Luan, Amodei, and
  Sutskever]{radford2019gpt}
Radford, A., Wu, J., Child, R., Luan, D., Amodei, D., and Sutskever, I.
\newblock Language models are unsupervised multitask learners.
\newblock 2019.

\bibitem[Raganato et~al.(2020)Raganato, Scherrer, and
  Tiedemann]{raganato2020fixed}
Raganato, A., Scherrer, Y., and Tiedemann, J.
\newblock Fixed encoder self-attention patterns in transformer-based machine
  translation.
\newblock \emph{arXiv preprint arXiv:2002.10260}, 2020.

\bibitem[Rajpurkar et~al.(2016)Rajpurkar, Zhang, Lopyrev, and
  Liang]{rajpurkar2016squad}
Rajpurkar, P., Zhang, J., Lopyrev, K., and Liang, P.
\newblock {SQuAD}: 100,000+ questions for machine comprehension of text.
\newblock \emph{arXiv preprint arXiv:1606.05250}, 2016.

\bibitem[Sanh et~al.(2019)Sanh, Debut, Chaumond, and Wolf]{sanh2019distilbert}
Sanh, V., Debut, L., Chaumond, J., and Wolf, T.
\newblock Distilbert, a distilled version of bert: smaller, faster, cheaper and
  lighter.
\newblock \emph{arXiv preprint arXiv:1910.01108}, 2019.

\bibitem[Sanh et~al.(2020)Sanh, Wolf, and Rush]{sanh2020movement}
Sanh, V., Wolf, T., and Rush, A.
\newblock Movement pruning: Adaptive sparsity by fine-tuning.
\newblock \emph{Advances in Neural Information Processing Systems},
  33:\penalty0 20378--20389, 2020.

\bibitem[Shen et~al.(2020)Shen, Dong, Ye, Ma, Yao, Gholami, Mahoney, and
  Keutzer]{shen2020q}
Shen, S., Dong, Z., Ye, J., Ma, L., Yao, Z., Gholami, A., Mahoney, M.~W., and
  Keutzer, K.
\newblock {Q-BERT}: Hessian based ultra low precision quantization of bert.
\newblock In \emph{AAAI}, pp.\  8815--8821, 2020.

\bibitem[Socher et~al.(2013)Socher, Perelygin, Wu, Chuang, Manning, Ng, and
  Potts]{socher2013recursive}
Socher, R., Perelygin, A., Wu, J., Chuang, J., Manning, C.~D., Ng, A.~Y., and
  Potts, C.
\newblock Recursive deep models for semantic compositionality over a sentiment
  treebank.
\newblock In \emph{Proceedings of the 2013 conference on empirical methods in
  natural language processing}, pp.\  1631--1642, 2013.

\bibitem[Sun et~al.(2019)Sun, Cheng, Gan, and Liu]{sun2019patient}
Sun, S., Cheng, Y., Gan, Z., and Liu, J.
\newblock Patient knowledge distillation for bert model compression.
\newblock \emph{arXiv preprint arXiv:1908.09355}, 2019.

\bibitem[Sun et~al.(2020{\natexlab{a}})Sun, Wang, Chen, Ni, Agrawal, Cui,
  Venkataramani, El~Maghraoui, Srinivasan, and Gopalakrishnan]{sun2020ultra}
Sun, X., Wang, N., Chen, C.-Y., Ni, J., Agrawal, A., Cui, X., Venkataramani,
  S., El~Maghraoui, K., Srinivasan, V.~V., and Gopalakrishnan, K.
\newblock Ultra-low precision 4-bit training of deep neural networks.
\newblock \emph{Advances in Neural Information Processing Systems},
  33:\penalty0 1796--1807, 2020{\natexlab{a}}.

\bibitem[Sun et~al.(2020{\natexlab{b}})Sun, Yu, Song, Liu, Yang, and
  Zhou]{sun2020mobilebert}
Sun, Z., Yu, H., Song, X., Liu, R., Yang, Y., and Zhou, D.
\newblock Mobilebert: a compact task-agnostic bert for resource-limited
  devices.
\newblock \emph{arXiv preprint arXiv:2004.02984}, 2020{\natexlab{b}}.

\bibitem[Tang et~al.(2022)Tang, Zhang, Liu, Zhu, and Kang]{tang2022mkq}
Tang, H., Zhang, X., Liu, K., Zhu, J., and Kang, Z.
\newblock Mkq-bert: Quantized bert with 4-bits weights and activations.
\newblock \emph{arXiv preprint arXiv:2203.13483}, 2022.

\bibitem[Tenney et~al.(2019)Tenney, Das, and Pavlick]{tenney2019bert}
Tenney, I., Das, D., and Pavlick, E.
\newblock Bert rediscovers the classical nlp pipeline.
\newblock \emph{arXiv:1905.05950}, 2019.

\bibitem[Vaswani et~al.(2017)Vaswani, Shazeer, Parmar, Uszkoreit, Jones, Gomez,
  Kaiser, and Polosukhin]{vaswani2017attention}
Vaswani, A., Shazeer, N., Parmar, N., Uszkoreit, J., Jones, L., Gomez, A.~N.,
  Kaiser, {\L}., and Polosukhin, I.
\newblock Attention is all you need.
\newblock In \emph{Advances in neural information processing systems}, pp.\
  5998--6008, 2017.

\bibitem[Wang et~al.(2020)Wang, Wei, Dong, Bao, Yang, and Zhou]{wang2020minilm}
Wang, W., Wei, F., Dong, L., Bao, H., Yang, N., and Zhou, M.
\newblock Minilm: Deep self-attention distillation for task-agnostic
  compression of pre-trained transformers.
\newblock \emph{arXiv preprint arXiv:2002.10957}, 2020.

\bibitem[Warstadt et~al.(2018)Warstadt, Singh, and Bowman]{warstadt2018neural}
Warstadt, A., Singh, A., and Bowman, S.~R.
\newblock Neural network acceptability judgments.
\newblock \emph{arXiv preprint arXiv:1805.12471}, 2018.

\bibitem[Williams et~al.(2017)Williams, Nangia, and Bowman]{williams2017broad}
Williams, A., Nangia, N., and Bowman, S.~R.
\newblock A broad-coverage challenge corpus for sentence understanding through
  inference.
\newblock \emph{arXiv preprint arXiv:1704.05426}, 2017.

\bibitem[Wu et~al.(2022)Wu, Yao, Zhang, Li, and He]{wu2022extreme}
Wu, X., Yao, Z., Zhang, M., Li, C., and He, Y.
\newblock Extreme compression for pre-trained transformers made simple and
  efficient.
\newblock \emph{arXiv preprint arXiv:2206.01859}, 2022.

\bibitem[Xiao et~al.(2022)Xiao, Lin, Seznec, Demouth, and
  Han]{xiao2022smoothquant}
Xiao, G., Lin, J., Seznec, M., Demouth, J., and Han, S.
\newblock Smoothquant: Accurate and efficient post-training quantization for
  large language models.
\newblock \emph{arXiv preprint arXiv:2211.10438}, 2022.

\bibitem[Xiong et~al.(2020)Xiong, Yang, He, Zheng, Zheng, Xing, Zhang, Lan,
  Wang, and Liu]{xiong2020layer}
Xiong, R., Yang, Y., He, D., Zheng, K., Zheng, S., Xing, C., Zhang, H., Lan,
  Y., Wang, L., and Liu, T.
\newblock On layer normalization in the transformer architecture.
\newblock In \emph{International Conference on Machine Learning}, pp.\
  10524--10533. PMLR, 2020.

\bibitem[{Yao} et~al.(2021){Yao}, {Wu}, {Ma}, {Shen}, {Keutzer}, {Mahoney}, and
  {He}]{yao2021mlpruning}
{Yao}, Z., {Wu}, X., {Ma}, L., {Shen}, S., {Keutzer}, K., {Mahoney}, M.~W., and
  {He}, Y.
\newblock {LEAP: Learnable Pruning for Transformer-based Models}.
\newblock \emph{arXiv e-prints}, art. arXiv:2105.14636, May 2021.

\bibitem[Yao et~al.(2022)Yao, Aminabadi, Zhang, Wu, Li, and
  He]{yao2022zeroquant}
Yao, Z., Aminabadi, R.~Y., Zhang, M., Wu, X., Li, C., and He, Y.
\newblock Zeroquant: Efficient and affordable post-training quantization for
  large-scale transformers.
\newblock In Oh, A.~H., Agarwal, A., Belgrave, D., and Cho, K. (eds.),
  \emph{Advances in Neural Information Processing Systems}, 2022.
\newblock URL \url{https://openreview.net/forum?id=f-fVCElZ-G1}.

\bibitem[Yao et~al.(2023)Yao, Li, Wu, Youn, and He]{yao2023comprehensive}
Yao, Z., Li, C., Wu, X., Youn, S., and He, Y.
\newblock A comprehensive study on post-training quantization for large
  language models.
\newblock \emph{arXiv preprint arXiv:2303.08302}, 2023.

\end{thebibliography}
}

\clearpage
\onecolumn
\appendix

\section{Quantization Algorithms}\label{sec:method-app}
\textbf{Weight Quantization.} To quantize a weight matrix $\rmW \in \sR^{d_{in}\times d_{out}}$ in a model, we apply the \textit{group-wise} quantization method proposed in~\citet{shen2020q, yao2022zeroquant}. That is, we vectorize  $\text{Vectorize}(\rmW) \in \sR^{d}$ ($d=d_{in} d_{out}$) and participate the weight into $g$ groups, and each group is quantized separately. The finer the quantization we use (larger $g$), the smaller the approximation error is between the weight matrix and the 4-bit counterpart. 
The largest group number we apply here is $d_{in}$, i.e., a row-wise weight quantization for best GPU utilization, 
 
\textbf{Activation Quantization.} Different from the static weight parameter during inference, the activation is dynamic. In order to achieve the best latency reduction \citep{gholami2021survey}, the static quantization method calibrates $S$ using training
data and fixes $S$ during inference. However, this also limits the quantization representation for activation as discussed in \cite{yao2022zeroquant}. Thus, we adopt a finer-grained token-wise
dynamic quantization and use the min/max range for each token.
\counterwithin{figure}{section}
\counterwithin{table}{section}

\section{Additional Experimental Details and Results}
\label{sec:experiment-setup}

\subsection{Experimental Details for \sref{sec:results}}
All experiments are performed on V100 GPU and the training strategy is Quantization-aware Training (QAT) with Knowledge Distillation (KD).
For BERT-models, the maximum sequence length for MNLI/QQP is set to 128 with a batch-size of 64 for base and 32 for large. Each independent experiment is ran on a single GPU with a fixed random seed 42.  We recorded their validation performances over the training every 1000 iterations and report the best validation value.  For BART-type models, we follows closely with \cite{li2022dq} with a slightly different hyper-parameters as shown in \tref{table:main-results-hp}. Each independent experiment is ran on 2 GPUs for base models and 4 GPUs for large model. As for GPT2-type models, we  applied the pretrained hugging-face models with maximum length 1024 and a batch-size of 8 with 4 GPUs for an independent run. See  \tref{table:main-results-hp} for the hyper-parameters search and  we will open sources the codes and the configurations.

\begin{figure}[ht]
    \centering
    \includegraphics[width=0.32\textwidth]{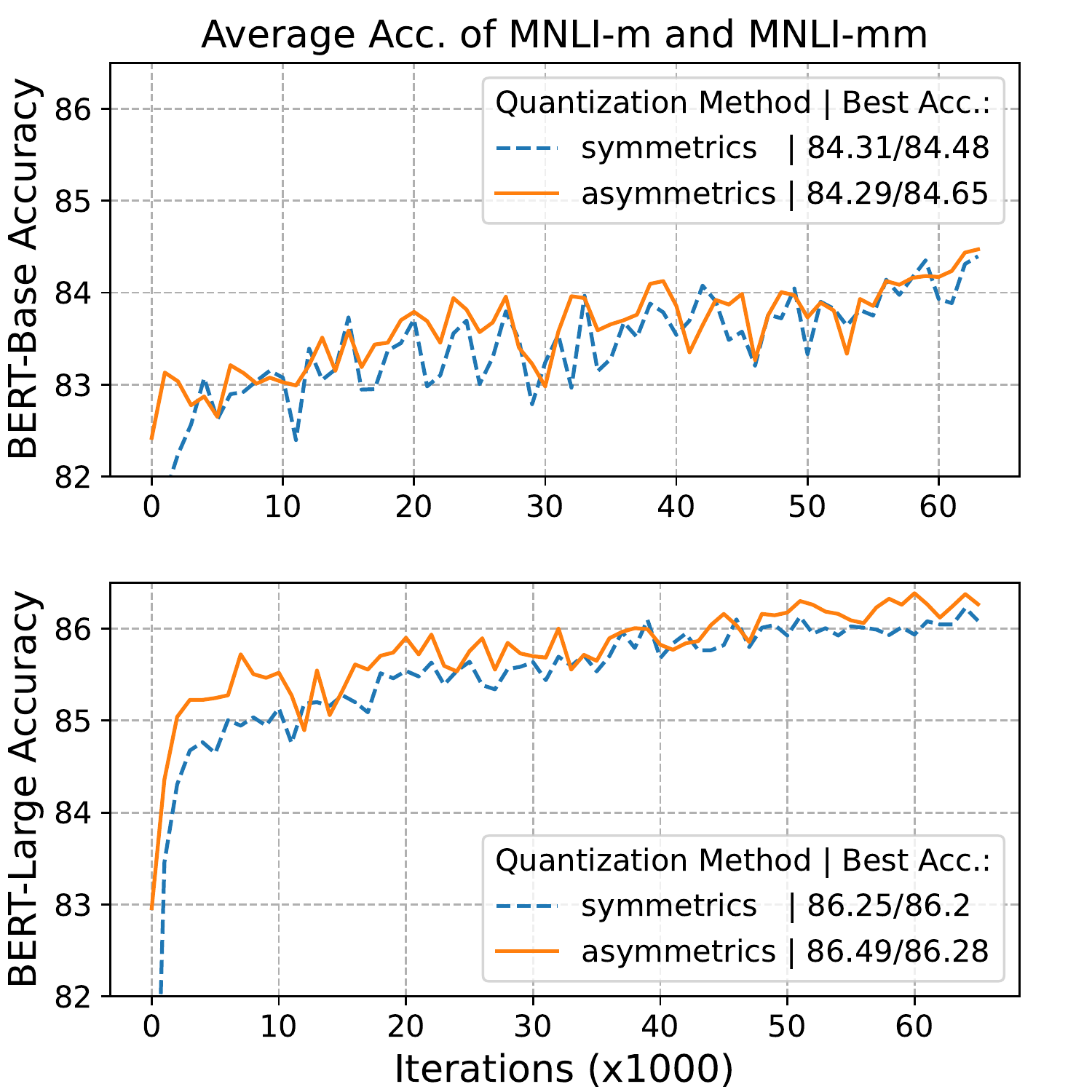}
    \includegraphics[width=0.32\textwidth]{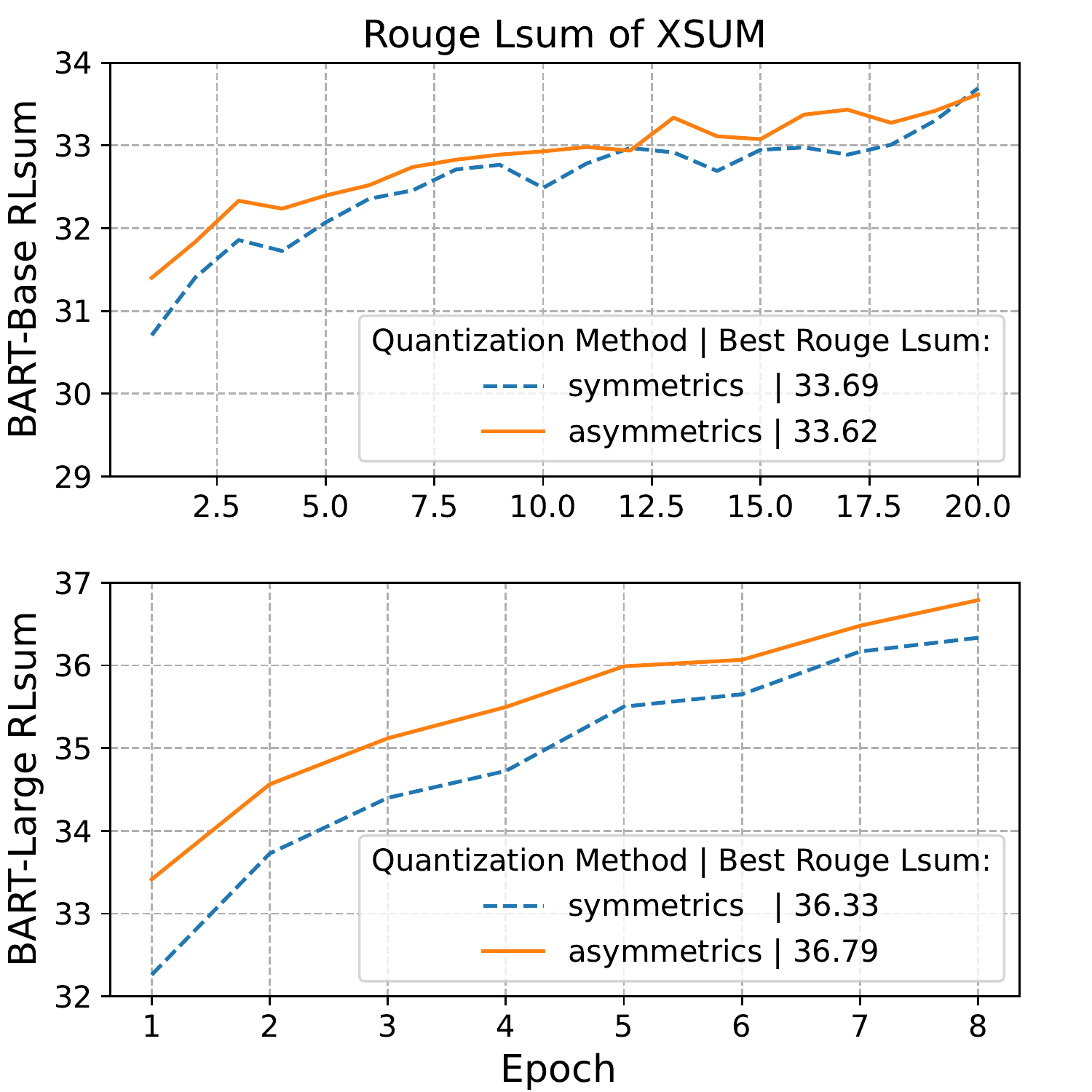}
    \includegraphics[width=0.32\textwidth]{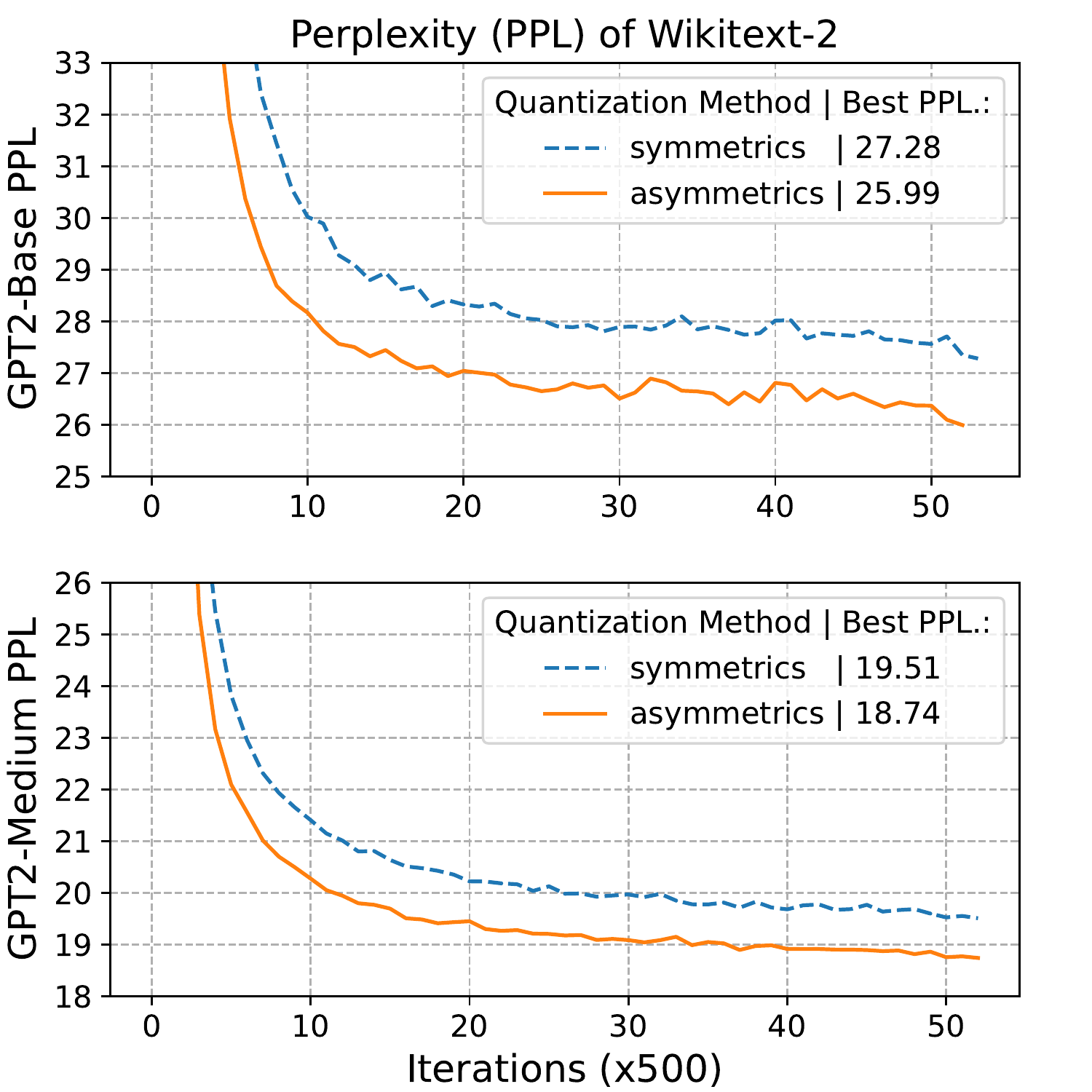}
\caption{The performance of w4a4 during the quantization-aware trainig with KD over the validation dataset for BERT (left), BART (middle), GPT (right) models, respectively with metrics: Accuracy (Acc., higher is better), Rouge Lsum (RLsum, higher is better), and perplexity (PPL, lower is better). Molde sizes in the top row are smaller than those in the bottom row.}\label{fig:main-results}
\end{figure}

\textbf{Asymmetric and Symmetric Quantization.} To give better understand on the difference of asymmetric and symmetric quantization, we plot in \fref{fig:main-results} for the iterative performance over the validation datasets during the quantized-aware training.  The orange curves always sits on top of the blue dash line, proving the assymetric quantization is better than symmetric. Furthermore, \fref{fig:main-results} shows that (1) The gaps between symemtric and assymetric quantization appears more obvious as the model size increase from base (the first row) to large/medium (the second row), which indicates the importance of asymmetric quantization for large models; (2) While the benefits of asymmetric method (over the symmetric one) could become marginal from the beginning of the training to the end, it appears that is only the case for BERT and BART not for GPT. 

\begin{table}[ht]
\caption{The hyper-parameters we tuned for the results in \tref{table:main-results}. The entry with single choice means we only use the default value. For the entry with multiple choice, we bold the one that gives the best performance. In the table,  Att with $\star$ (Att{$^\star$}) means the attention scores that is not normalized, and Att is a normalized version (note that the default output from Huggingface library is a normalized venison of attention scores).
 }\label{table:main-results-hp}
\begin{adjustbox}{width=0.999\linewidth}
\centering
\begin{tabular}{l|cc|cc|cc}
\toprule
Models     & \multicolumn{2}{c|}{BERT}                         & \multicolumn{2}{c|}{BART}                                  & \multicolumn{2}{c}{GPT }                                         \\
Size      & {Base} &  {Large}    & {Base} &  {Large}     & {Base} & {Medium}\\ \midrule  
Dropout &   \multicolumn{2}{c|}{0.1 (default)}         & \multicolumn{2}{c|}{0.1 (default)} &  \multicolumn{2}{c}{ \{0, \textbf{0.05}, 0.1\}}     \\
Clip Values & \multicolumn{2}{c|}{\{\textbf{[-5.0, 5.0]}, [-$\infty$, +$\infty$]\}}  & \{\textbf{[-1,1]},[-2.5,2.5]\}&[-2.5, 2.5] &\multicolumn{2}{c}{\{[-0.5,0.5],\textbf{[-1, 1]}, [-2.5, 2.5]\}}                                       \\
Loss Terms &  \multicolumn{2}{c|}{Logit/Att$^\star$/Rep (default)}  &  \multicolumn{2}{c|}{ClsLoss/Logit/Att$^\star$/Rep} & \multicolumn{2}{c}{\{None,\textbf{ClsLoss}\}+\{Att,\textbf{Att$^\star$}\}+Logit/Rep}                                        \\
Epoch &  \{3, \textbf{9}\}  & \{3, \textbf{5}\} &  20 & 8 & \{30, 60, \textbf{90}\} &  \{15, 30, \textbf{45}\}\\
batch-size &  64  & 32 & 16 & 8 & 8 & 8 \\
Learning Rate &  \{5e-5, \textbf{1e-4}\}  & \{2e-5, \textbf{5e-5}\} & \{2e-5, \textbf{5e-5}\} & { \{2e-5, \textbf{5e-5}\}} & \{{5e-5}, \textbf{1e-4}, 5e-4\} & \{{5e-5}, \textbf{1e-4}\}\\
 \bottomrule
\end{tabular}
\end{adjustbox}
\end{table}

\subsection{MKQ-BERT Results}\label{sec:mkq-results}

Table~\ref{table:mkq} shows the latency for a single BERT-base layer reported by MKQ-BERT\cite{tang2022mkq} Table 2, compared to FasterTransformer\cite{fastertransformer}. 
We can see that both the FP32 and INT8 results are off by more than an order of magnitude. Due to the lack of implementation details described in the MKQ-BERT paper (no open-sourced code), we cannot further identify the issue.
\begin{table}[h]
\begin{center}
\caption{End-to-end inference time (ms) for running one layer in BERT-base model with different batch size and sequence length on NVIDIA T4 GPUs. Column 2 to 4 are numbers taken from \cite{tang2022mkq}.
FasterTransformer(FT) requires sequence length to be multiple of 32, thus the inputs in the parenthesis are used to run FasterTransformer.} 
\begin{tabular}{|c|ccc|cc|}
\hline
 Batch Size-Seq. Length &  MKQ-fp32 & MKQ-int8 & MKQ-in4 & FT-fp32 & FT-int8  \\
 \hline
16-440 (16-448) & 1.38	& 0.2131 & 	0.1605 & 25.62 & 5.1 \\
16-537 (16-544) & 1.845	& 0.2457 & 	0.1793 & 34.25 & 6.61 \\
16-681 (16-704) & 2.69	& 0.2609 & 	0.1965 &  46.54 & 9.39 \\
\hline
\end{tabular}\label{table:mkq}
\end{center}
\end{table}

\subsection{Sensitivity of Activation Quantization for GPT2}
In this section, we study how sensitive the model quality to activation quantization. We relax the INT4 activation to be INT8 or back to FP32, and follows the same QAT recipe as W4A4. We plot average perplexity with respect to the training iteration in \fref{figure:gap-activation} (left) as well as the position perplexity at training iteration 34000 in \fref{figure:gap-activation} (right). We see that  although using QAT with KD,  w4a8 (green) can be better than W4A4 but still far away from teacher's quality.  Only 4-bit weight quantization (w4 only, red curve) can almost match the teacher's quality (blue), which indicates that autoregressive generation using GPT models is highly sensitive to activation quantization. It is interesting to notice that the red curve in \fref{figure:gap-activation} (left) already flatten at the beginning of the training, which means PTQ method could be possible for weight only quantization, this aligns with observation in \citet{dettmers2022case}.

\subsection{More Experiments on Composing Pruning and INT4}
Besides the MNLI/QQP mentioned in \sref{sec:results-bert}, we  include  the following GLUE tasks for the W4A4 quantization:
MRPC~\citep{dolan2005automatically},
STS-B~\citep{cer2017semeval},
SST-2~\citep{socher2013recursive}, QNLI~\citep{rajpurkar2016squad}, QQP~\citep{iyer2017first}, MNLI~\citep{williams2017broad}, CoLA~\citep{warstadt2018neural}, RTE~\citep{dagan2013recognizing}).  The maximum sequence length is set to 64 for CoLA/SST-2, and 128 for the rest sequence pair tasks.

\begin{table}[ht]
\caption{Comparison between static and iterative pruning methods on top of w4a4 models. Here the 50\% sparsity  is semi-struture pruning with Pair-(2:4). We applied data augmentation for the smaller datasets and used the long training epochs (Budget-C) shown in Table \ref{table:budgets}. The learning rate is fixed with 1e-4. Note the results for MNLI and QQP are different from \tref{table:mnli} is due to the teacher models. }\label{table:compre-pruning}

\begin{adjustbox}{width=0.99\linewidth}
\begin{tabular}{lccccccccccccccc}
\toprule
Model  & Pruning Method&  CoLA  & MNLI-m/-mm  &  MRPC      &  QNLI  &    QQP      & RTE    & SST-2 &   STS-B    & Avg.   \\
    & & Mcc     &   Acc/Acc   & F1/Acc     &  Acc   &  F1/Acc     & Acc    &  Acc  & Pear/Spea  &    all   \\   \midrule  
\multicolumn{2}{l}{BERT$_{\text{base}}$ (teacher) }         &  59.7  &  84.9/85.6   &  90.6/86.3  &  92.1  &  88.6/91.5   & 72.2   & 93.2  & 90.1/89.6 & 83.95 \\\midrule
\multirow{2}{*}{w4a4+ 50\% sparsity} &Static  (weight)                             & 57.4  & 84.4/84.8  & 90.8/86.5  & 91.4  & 88.3/91.4  & 73.3  & 93.3  & 89.5/89.2  & 83.56 \\       
 & Iterative  (gradient)    & 58.1  & 84.3/84.9  & 90.9/86.8  & 91.4  & 88.4/91.4  & 72.9  & 93.3  & 89.4/89.0  & 83.61  \\ 
 \bottomrule
\end{tabular}
\end{adjustbox}
\end{table}

\begin{table}[h]
\centering
\caption{Training budgets for the GLUE tasks.}
\label{table:budgets}
\begin{adjustbox}{width=0.45\linewidth}
\begin{tabular}{lcccc}
\toprule
\multirow{2}{*}{Dataset} & Data  &\multicolumn{3}{c}{Training epochs:} \\
                     & Aug.  &  Budget-A  &  Budget-B  &  Budget-C  \\\midrule
    QQP/MNLI & no &  3  & 9  & 18 or 36 \\
    QNLI  & yes   &  1  & 3  &  6 or 9 \\
  SST-2/STS-B/RTE & yes &  1  & 3  &  12 \\
    CoLA/MRPC  & yes  &  1  & 3  &  12 or 18 \\
\bottomrule
\end{tabular}
\end{adjustbox}
\end{table}

\begin{table}[ht]
\caption{Results with data augmentation.  \textbf{W4A4} with Budget-A}\label{table:da-1bit-app}
\begin{adjustbox}{width=0.99\linewidth}
\centering
\begin{tabular}{lccccccccccccccc }
\toprule
Cost& learning    &  CoLA  & MNLI-m/-mm  &  MRPC      &  QNLI  &    QQP      & RTE    & SST-2 &   STS-B    & Avg. &Avg. w/o     \\
  &  rate  & Mcc     &   Acc/Acc   & F1/Acc     &  Acc   &  F1/Acc     & Acc    &  Acc  & Pear/Spea  &    all  &  CoLA   \\   \midrule  
 \multicolumn{2}{c}{BERT$_{\text{large}}$ (fp32) }         &  63.4  &  85.4/85.4   &  91.6/88.0  &  92.2  &  88.4/91.05   & 74.00   & 93.6  & 90.3/90.1 & 84.81 & 87.50\\\cdashline{1-13}
\multirow{4}{*}{Budget-A} 
           & 1e-05  & 62.1  & 85.6/85.3  & 91.2/87.3  & 92.5  & 88.3/91.3  & 69.7  & 93.7  & 90.3/90.1  & 84.20 & 86.96 \\ 
                                       & 5e-05   & 64.6  & 85.6/85.2  & 91.0/87.0  & 92.5  & 87.7/90.9  & 72.2  & 93.8  & 90.7/90.4  & 84.72 & 87.24 \\ 
                                       & 0.0001   & 61.7  & 85.8/85.4  & 90.9/87.0  & 91.9  & 88.5/91.5  & 75.5  & 93.8  & 90.6/90.3  & 84.80 & 87.69 \\ \cdashline{2-13}
                               &  Best (above)  & 64.6  & 85.8/85.4  & 91.2/87.3  & 92.5  & 88.5/91.5  & 75.5  & 93.8  & 90.7/90.4  & 85.23 & 87.81 \\ \midrule
 \multicolumn{2}{c}{BERT$_{\text{base}}$ (fp32) }         &  59.7  &  84.9/85.6   &  90.6/86.3  &  92.1  &  88.6/91.5   & 72.2   & 93.2  & 90.1/89.6 & 83.95 & 86.98\\\cdashline{1-13}
\multirow{5}{*}{ Budget-A} 
                                       & 2e-05  & 59.1  & 84.5/85.0  & 91.2/87.0  & 91.7  & 88.3/91.3  & 73.6  & 93.7  & 89.8/89.4  & 83.97 & 87.08 \\ 
                                       & 5e-05   & 60.5  & 84.6/85.0  & 90.4/85.8  & 91.6  & 88.4/91.4  & 71.5  & 93.3  & 90.0/89.6  & 83.74 & 86.65 \\ 
                                       & 0.0001   & 59.9  & 84.8/85.2  & 90.4/85.8  & 92.0  & 88.3/91.4  & 72.6  & 93.5  & 89.9/89.6  & 83.90 & 86.90 \\ 
                                       \cdashline{2-13}
                                       &  Best (above)  & 60.5  & 84.8/85.2  & 91.2/87.0  & 92.0  & 88.4/91.4  & 73.6  & 93.7  & 90.0/89.6  & 84.24 & 87.21 \\
 \bottomrule
\end{tabular}
\end{adjustbox}
\end{table}

 \begin{figure}[H]
 \centering
 \includegraphics[width=0.595\textwidth]{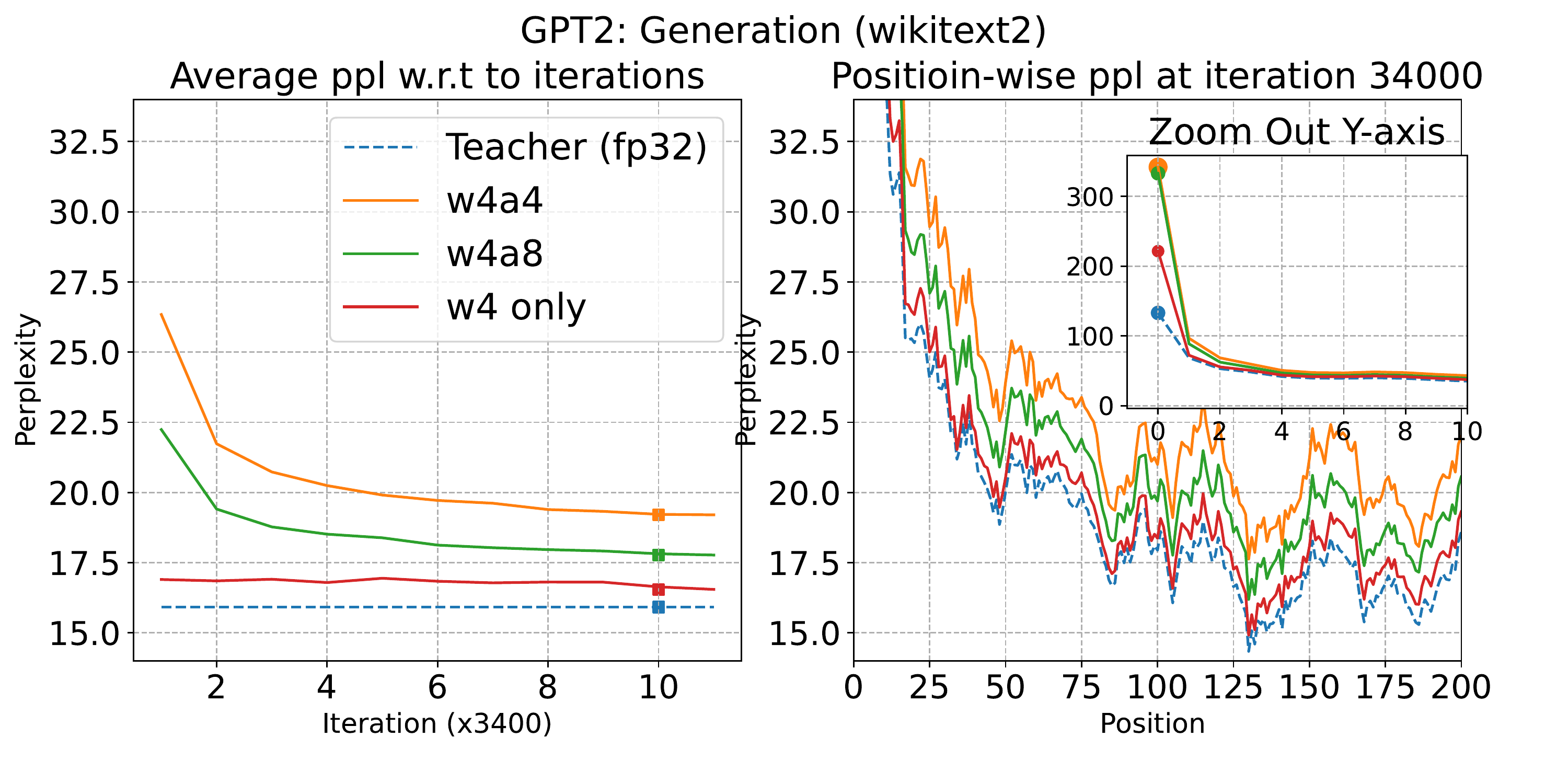}
 \includegraphics[width=0.595\textwidth]{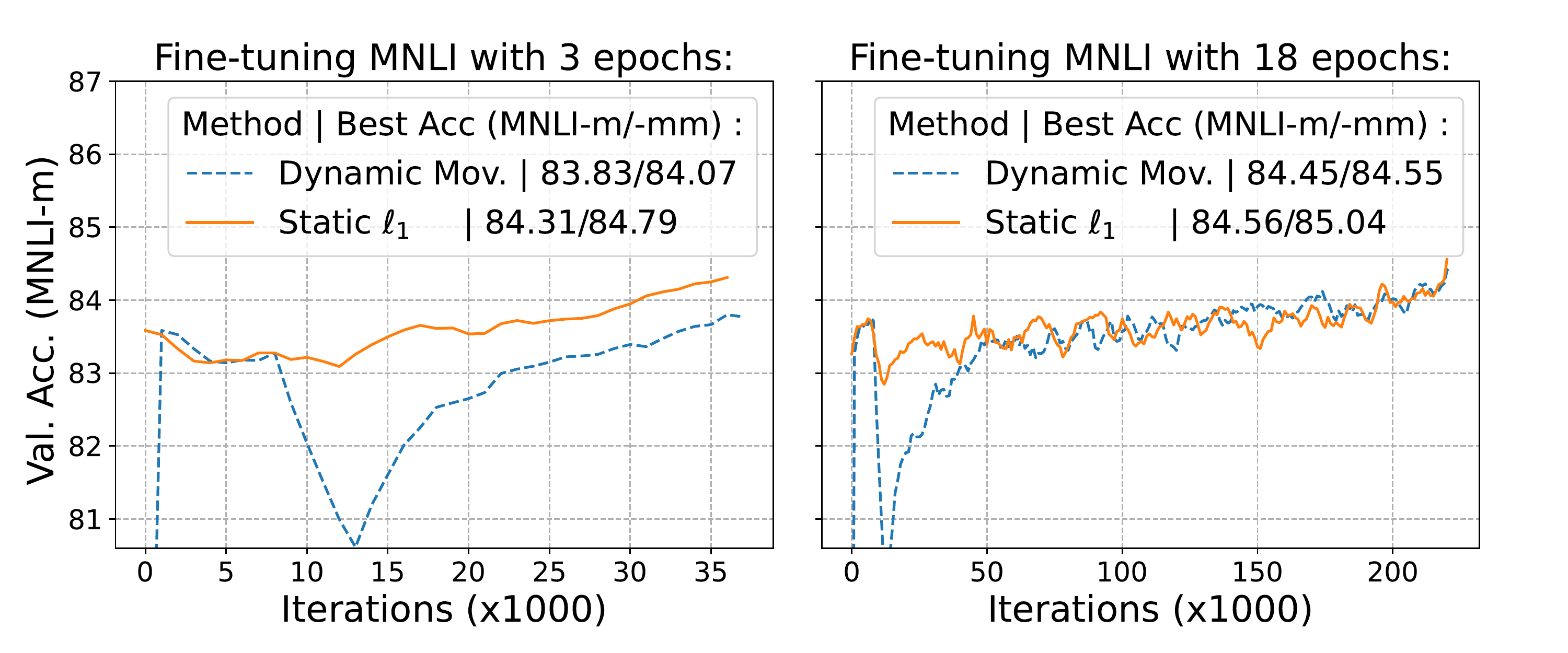}
  \caption{The top two figures are for GPT2 quantization. The bottom two figures are for the comparison between movement and $\ell_1$ pruning with QAT.}\label{fig:compare}
 \end{figure}
    \textbf{Static pruning or iterative movement pruning.} Now that we decide to apply prune and then quantize (P$=>$Q) algorithm, one may wonder if the $\ell_1$ pruning method used above is the best pruning algorithm.  Recent advancement on pruning methods suggests that Movement Pruning with iterative pruning threshold~\citep{sanh2020movement,lagunas2021block} has been proven to be effective in transfer learning for languages models. That is, during the iterative pruning, the mask will be updated  and  determined by the gradients of the weight instead of the value of the weight. Previous works only work on pruning only, here we investigate on whether it works well with quantized models and layerwise KD. 
    
    The results are shown in \fref{fig:compare} or \tref{table:compre-pruning}. We see that   $\ell_1$ is consistently better  under long or short training epochs, although the gap between the two methods can be reduced with sufficient iterations. As the iterative pruning based on the gradient of weight matrices requires to update the masks dynamically, the computation complexity/time is much higher than that using the static masks under the same training iterations. Thus, this finding indicates that static pruning is sufficient when applying KD for QAT.


\end{document}